\documentclass[twocolumn,journal]{IEEEtran}
\usepackage[T1]{fontenc}
\usepackage[latin9]{inputenc}
\usepackage{color}
\usepackage{amsmath}
\usepackage{amssymb}
\usepackage{stackrel}
\usepackage{graphicx}
\usepackage[unicode=true,
 bookmarks=true,bookmarksnumbered=true,bookmarksopen=true,bookmarksopenlevel=1,
 breaklinks=false,pdfborder={0 0 0},pdfborderstyle={},backref=false,colorlinks=false]
 {hyperref}
\hypersetup{pdftitle={Your Title},
 pdfauthor={Your Name},
 pdfpagelayout=OneColumn, pdfnewwindow=true, pdfstartview=XYZ, plainpages=false}

\makeatletter
\usepackage[caption=false,font=footnotesize]{subfig}
\usepackage{booktabs}
\usepackage{multirow}
\usepackage{stmaryrd}
\usepackage{color}
\usepackage{xcolor}

\@ifundefined{showcaptionsetup}{}{%
 \PassOptionsToPackage{caption=false}{subfig}}
\usepackage{subfig}
\makeatother

\begin{document}
\title{Fully Linear Graph Convolutional Networks for Semi-Supervised Learning
and Clustering}
\author{Yaoming~Cai,~\IEEEmembership{Student Member,~IEEE,} Zijia~Zhang,~Zhihua~Cai,~Xiaobo~Liu,~\IEEEmembership{Member,~IEEE},~Yao~Ding,~and~Pedram
Ghamisi \IEEEmembership{Senior Member,~IEEE}\thanks{This work was supported in part by the National Natural Science Foundation
of China (NSFC) under Grant 61773355 and Grant 61973285; in part by
the Fundamental Research Funds for National University, China University
of Geosciences (Wuhan), under Grant CUGGC03, Grant G1323541717, and
Grant 1910491T06; and in part by the National Nature Science Foundation
of Hubei Province under Grant 2018CFB528.}\thanks{Yaoming Cai, Zijia Zhang, and Zhihua Cai are with the School of Computer
Science, China University of Geosciences, Wuhan 430074, China (e-mail:
caiyaom@cug.edu.cn; zhangzijia@cug.edu.cn; zhcai@cug.edu.cn).}\thanks{Xiaobo Liu is with the School of Automation, China University of Geo-
sciences, Wuhan 430074, China, and also with the Hubei Key Laboratory
of Advanced Control and Intelligent Automation for Complex Systems,
China University of Geosciences, Wuhan 430074, China (e-mail: xbliu@cug.edu.cn).}\thanks{Ding Yao is with the Xi\textquoteright an Research Institute of High
Technology, Xi\textquoteright an 710000, China (e-mail: dingyao.88@outlook.com).}\thanks{Pedram Ghamisi is with the Helmholtz-Zentrum Dresden-Rossendorf (HZDR),
Helmholtz Institute Freiberg for Resource Technology, 09599 Freiberg,
Germany, and also with the Institute of Advanced Research in Artificial
Intelligence (IARAI), 1030 Vienna, Austria (e-mail: p.ghamisi@gmail.com).}}
\maketitle
\begin{abstract}
This paper presents FLGC, a simple yet effective fully linear graph
convolutional network for semi-supervised and unsupervised learning.
Instead of using gradient descent, we train FLGC based on computing
a global optimal closed-form solution with a decoupled procedure,
resulting in a generalized linear framework and making it easier to
implement, train, and apply. We show that (1) FLGC is powerful to
deal with both graph-structured data and regular data, (2) training
graph convolutional models with closed-form solutions improve computational
efficiency without degrading performance, and (3) FLGC acts as a natural
generalization of classic linear models in the non-Euclidean domain,
e.g., ridge regression and subspace clustering. Furthermore, we implement
a semi-supervised FLGC and an unsupervised FLGC by introducing an
initial residual strategy, enabling FLGC to aggregate long-range neighborhoods
and alleviate over-smoothing. We compare our semi-supervised and unsupervised
FLGCs against many state-of-the-art methods on a variety of classification
and clustering benchmarks, demonstrating that the proposed FLGC models
consistently outperform previous methods in terms of accuracy, robustness,
and learning efficiency. The core code of our FLGC is released at
https://github.com/AngryCai/FLGC.

\end{abstract}


\section{Introduction}

Graph neural network (GNN) has emerged as a powerful technique for
the representation learning of graphs \cite{GNN-Survey-ZHNAG-TKDE-2020,GNN-Review-Zhou-arXiv-2018}.
Current GNN models usually follow a neighborhood aggregation scheme,
where the feature representation of a node is updated by recursively
aggregating representation vectors of its neighbors \cite{GCN-Deeperinsights-Li-AAAI-2018,GNN-Review-Wu-TNNLS-2020}.
Benefited from the promising ability to model graph structural data,
traditional learning problems can be revisited from the perspective
of graph representation learning, ranging from recommendation \cite{Recommendation-gnn-WWW-19}
and computer vision \cite{GNN-CV-Monti-CVPR-17,HSI-Clustering-GCSC-CAI-TGRS-2020,YaoDing-Graph-JSTARS-2021},
to combinatorial optimization \cite{Combinatorial-Optimization-GNN-NIPS-19,Combinatorial-Optimization-GNN-Wilder-NIPS-19}.

In recent years, there is increasing attention for generalizing convolution
to the graph domain. The existing graph convolutional networks (GCNs)
often broadly categorize into spectral approaches and spatial approaches
\cite{GNN-Review-Zhou-arXiv-2018,GNN-Review-Wu-TNNLS-2020}. Spectral
models commonly use the spectral graph theory to design spectral filters,
such as ChebyNet \cite{ChebyNet-Deff-NIPS-16} and the vanilla GCN
\cite{GCN-kipf-ICLR-2017}. While spatial models define graph convolutions
based on a node\textquoteright s spatial relations, e.g., GraphSAGE
\cite{GraphSage-William-NIPS-17,YaoDing-Graph-GRSL-2021} and Graph
Attention networks (GAT) \cite{GAT-PETAR-ICLR-2018}. Due to the good
theoretical guarantee, spectral models have been extensively studied
by the recent mainstream. Despite their success, most of the state-of-the-art
(SOTA) GCN models are dominated by shallow and simple models because
of the over-smoothing problem \cite{GCN-Deeperinsights-Li-AAAI-2018,GCNII-Chen-ICML-2020}.
To circumvent the problem, many efforts have been paid to develop
deeper and more robust models. The proven techniques include residual
connection \cite{GCNII-Chen-ICML-2020}, randomly drop nodes \cite{GRNN-Feng-NIPS-20}
or edges \cite{DropEdge-ICLR-2019}, and data augmentation \cite{GAug-Zhao-AAAI-21},
etc. 

More recent attempts in simplifying GCN have collectively suggested
that a GCN model can be decoupled into two successive stages: parameter-free
neighborhood propagation and task-specific prediction. Following this
scheme, Wu \emph{et al.} \cite{GCN-simplifying-wu-ICML-2019} initially
proposed Simple Graph Convolution (SGC) \cite{GCN-simplifying-wu-ICML-2019}
by simplifying the vanilla GCN as a low-pass filter followed by a
linear classifier. While Approximate Personalized Propagation of Neural
Predictions (APPNP) \cite{APPNP-klicpera-ICLR-2019} swaps the two
stages and establishes a connection between GCN and the well-known
PageRank \cite{PageRank-page-1999}. These successful attempts signify
several helpful tendencies. First, simplified GCNs have a similar
capability to the elaborated ones in handling structural data. Second,
a linear GCN is as powerful as the nonlinear counterpart for most
graph scenarios. Nonetheless, fewer studies have managed to develop
a unified framework between traditional linear models and GCNs. Moreover,
gradient descent-based training of simplified GCNs often suffers from
local optimum, tedious hyper-parameters and training tricks. Hence,
it would result in a more simplified GCN if a globally optimal solution
in a closed form is provided.

Earlier linear models, such as ridge regression \cite{K-RR-AN-CVPR-2007}
and subspace clustering \cite{SSC-Elhamifar-TPAMI-2013}, have been
frequently applied in practice owing to their simplicity, efficiency,
and efficacy. However, these classic models essentially work in the
Euclidean domain, leading to insufficient handling of graph structured
data. Before GCNs emerged, graph Laplacian regularization (or manifold
regularization) \cite{Semi-LapRLS-LapSVM-JMLR-2006,HyperAE-CAI-GRSL-2021}
had been widely applied in various linear models to incorporate structural
information. This inspires a series of classic graph-regularized semi-supervised
approaches, e.g., Laplacian Regularized Least Squares (LapRLS) \cite{Semi-LapRLS-LapSVM-JMLR-2006},
and graph-regularized unsupervised approaches, e.g., Laplacian Regularized
Low-Rank Representation \cite{Laplacian-Regularized-LRR-TPAMI-YinM-2016}.
Despite being a useful technique, the Laplacian regularization encounters
three shortcomings: 1) it is typically dependent upon the Euclidean
domain and, thus, it is hard to directly generalize to real-word graph
data; 2) it merely considers the 1$^{st}$-order neighborhoods while
ignoring important long-range interdependency; 3) its additional regularization
coefficient needs to be appropriately adjusted.

In this paper, we propose a very simple and unified framework, referred
to as Fully Linear Graph Convolution (FLGC), for both semi-supervised
learning (SSL) and unsupervised learning (USL). Our goal is to further
simplify GCNs and generalize it to existing linear models, finally
enabling traditional models to explore graphs directly. Specifically,
we linearize GCN and then decouple it into neighborhood propagation
and prediction stages, resulting in a flexible framework. Such simplification
makes it possible to efficiently calculate a globally optimal solution
during training and also easy to incorporate with various previous
linear models. On the basis of the resulting framework, we further
present a semi-supervised FLGC for node classification problems and
an unsupervised FLGC for subspace clustering problems. To prevent
the over-smoothing issue, we introduce an initial residual in neighborhood
propagation to capture the long-range dependency across a graph.

To sum up, the main contributions of this paper are as follows:
\begin{enumerate}
\item We present a simple yet effective FLGC framework to deal with graph
data and regular data. The framework consists of a parameter-free
neighborhood propagation and a task-specific linear model with a closed-form
solution. The framework not only simplifies the training of existing
GCNs with a general but makes it easier to implement and apply in
practice.
\item We implement two FLGCs for semi-supervised node classification and
unsupervised clustering tasks, respectively. Also, we show that the
semi-supervised and unsupervised FLGCs act as a generalization of
classic ridge regression and subspace clustering in the non-Euclidean
domain. Such generalization enables traditional linear models to explore
graph structure without losing their original simplicity and efficacy.
\item We extend the personalized propagation scheme to balance the contribution
of structure and node features, which endows FLGC with the capability
of capturing long-range neighborhoods, thereby reducing the risk of
over-smoothing. 
\item We empirically show that FLGCs consistently outperform a large number
of previous SOTA approaches on both semi-supervised learning and clustering
tasks across real-world graph data and regular grid data. Such superiority
further offers a promising scheme to revisit traditional linear models
in a pure non-Euclidean domain.
\end{enumerate}
The rest of the paper is structured as follows. In Section \ref{sec:Notation-And-Concepts},
we briefly review the related concepts on recent GCN models and subspace
clustering. Section \ref{sec:FLGC} systematically introduces the
motivation, framework, formulation, and implementation of the proposed
FLGCs. Extensive qualitative and quantitative evaluations and comparisons
are presented in Section \ref{sec:Experiments-of-Traductive} and
Section \ref{sec:Experiments-of-Clustering}, followed by the conclusions.
Conclusions and future works are given in Section \ref{sec:Conclusions}.

\section{Notation and Concepts\label{sec:Notation-And-Concepts}}

\subsection{Notations}

In this paper, boldface lowercase italics symbols (e.g., $\boldsymbol{x}$),
boldface uppercase roman symbols (e.g., $\mathbf{X}$), regular italics
symbols (e.g., $N$), and calligraphy symbols (e.g., $\mathcal{S}$)
orderly denote vectors, matrices, scalars, and sets. A graph is represented
as $\mathcal{G}=(\mathcal{V},\mathcal{E},\mathbf{X})$, where $\mathcal{V}$
denotes the node set with $v_{i}\in\mathcal{V}$ and $\left|\mathcal{V}\right|=N$,
$\mathcal{E}$ indicates the edge set with $\left(v_{i},v_{i}\right)\in\mathcal{E}$,
and $\mathbf{X}\in\mathbb{R}^{N\times D}$ signifies the $D$-dimensional
node feature matrix of $\mathcal{G}$ while the corresponding $C$-class
one-hot target matrix is $\mathbf{Y}\in\mathbb{R}^{N\times C}$. We
define $\mathbf{A}\in\mathbb{R}^{N\times N}$ as the adjacency matrix
of $\mathcal{G}$ and the diagonal degree matrix as $\mathbf{D}\in\mathbb{R}^{N\times N}$,
where\textcolor{blue}{{} }$D_{ii}=\sum_{j}A_{ij}$. The graph Laplacian
matrix is defined as $\mathbf{L}=\mathbf{D}-\mathbf{A}$, and its
normalized version is given by $\mathbf{L}_{sym}=\mathbf{D}^{-1/2}\mathbf{L}\mathbf{D}^{-1/2}$,
which has an eigendecomposition of $\mathbf{U}\boldsymbol{\Lambda}\mathbf{U}^{T}$.
Here, $\boldsymbol{\Lambda}$ is a diagonal matrix of the eigenvalues
of $\mathbf{L}$, and $\mathbf{U}$ is a unitary matrix consisting
of the eigenvectors of $\mathbf{L}$. Bedsides, $\mathbf{X}^{T}$
denotes the transpose of matrix $\mathbf{X}$ and $\mathbf{I}_{N}$
denotes an identity matrix with the size of $N$. The trace operation
and the Frobenius norm of matrix $\mathbf{X}$ are defined as $\mathrm{tr}\left(\mathbf{X}\right)$
and $\mathbf{\left\Vert X\right\Vert }_{F}=\sqrt{\mathrm{tr}\left(\mathbf{X}^{T}\mathbf{X}\right)}$,
respectively.

\subsection{Vanilla GCN}

We refer to the GCN model proposed by Kipf \emph{et al}. \cite{GCN-kipf-ICLR-2017}
as the vanilla GCN because of its great success and numerous followers.
The vanilla GCN suggests that the graph convolution operation can
be approximated by the $K$-th order polynomial of Laplacians, i.e.,
\begin{equation}
\mathbf{U}g_{\theta}\mathbf{U}^{T}\boldsymbol{x}\approx\mathbf{U}\left(\stackrel[i=0]{K}{\sum}\theta_{i}\boldsymbol{\Lambda}^{\left(i\right)}\right)\mathbf{U}^{T}\boldsymbol{x}=\left(\stackrel[i=0]{K}{\sum}\theta_{i}\mathbf{L}^{\left(i\right)}\right)\boldsymbol{x},\label{eq:poly-gcn}
\end{equation}
where $g_{\theta}=diag\left(\theta\right)$ is a graph filter parameterized
by $\theta\in\mathbb{R}^{N}$ in the Fourier domain and $\theta_{i}$
denotes the polynomial coefficient. The vanilla GCN \cite{GCN-kipf-ICLR-2017}
adopts two crucial strategies to simplify and enhance Eq. \eqref{eq:poly-gcn}.
First, it uses a 1$^{st}$-order polynomial with settings of $\theta_{0}=2\theta$
and $\theta_{1}=-\theta$ to approximate Eq. \eqref{eq:poly-gcn},
resulting in a simplified convolution operation, i.e., $g_{\theta}*\boldsymbol{x}=\theta\left(\mathbf{I}+\mathbf{D}^{-1/2}\mathbf{A}\mathbf{D}^{-1/2}\right)\boldsymbol{x}$.
Second, it introduces a renormalization trick to guarantee its stability.
Specifically, the trick can be expressed as
\begin{equation}
\mathbf{P}=\left(\mathbf{D}+\mathbf{I}\right)^{-1/2}\left(\mathbf{A}+\mathbf{I}\right)\mathbf{\left(\mathbf{D}+\mathbf{I}\right)}^{-1/2}=\tilde{\mathbf{D}}^{-1/2}\mathbf{\tilde{A}}\mathbf{\tilde{D}}^{-1/2}.
\end{equation}
We call $\mathbf{P}$ as propagation matrix. As a result, in analogy
to convolutional neural networks (CNN) \cite{CNN-overview-GuJX-PR-2018,BS-Net-TGRS-Cai},
a general and layer-wise graph convolution propagation rule can be
defined by
\begin{equation}
\mathbf{\mathbf{X}^{\left(\ell+1\right)}}=\mathrm{ReLU}\left(\mathbf{P}\mathbf{X}^{\left(\ell\right)}\mathbf{W}^{\left(\ell\right)}\right).
\end{equation}
Here, $\mathbf{X}^{\left(\ell\right)}$ is the $\ell$-th layer's
graph embedding ($\mathbf{X}^{\left(1\right)}=\mathbf{X}$) and $\mathbf{W}^{\left(\ell\right)}$
is a trainable parameters matrix. However, many works \cite{GNN-Review-Zhou-arXiv-2018,GCN-Deeperinsights-Li-AAAI-2018,GCN-simplifying-wu-ICML-2019,APPNP-klicpera-ICLR-2019,GNN-Review-Wu-TNNLS-2020,GCNII-Chen-ICML-2020}
have demonstrated that stacking multiple GCN layers will lead to the
over-smoothing effect, that is to say, all vertices will converge
to the same value. Thus, the vanilla GCN usually adopts a shallow
architecture, e.g., two GCN layers.

\subsection{SGC}

SGC \cite{GCN-simplifying-wu-ICML-2019} removes nonlinear activations
in the vanilla GCN and collapses all trainable weights into a single
matrix. This enables it to raise the repeated multiplication of the
normalized adjacency matrix $\mathbf{P}$ as a $K$-th power of a
matrix, i.e., 
\begin{equation}
\begin{aligned}\tilde{\mathbf{Y}} & =\mathrm{softmax}\left(\underset{K}{\underbrace{\mathbf{P}\mathbf{P}\cdots\mathbf{P}}}\mathbf{\mathbf{X}}\underset{K}{\underbrace{\mathbf{W}_{1}\mathbf{W}_{2}\cdots\mathbf{W}_{K}}}\right)\\
 & =\mathrm{softmax}\left(\mathbf{P}^{K}\mathbf{\mathbf{X}}\mathbf{W}\right)
\end{aligned}
.
\end{equation}
Furthermore, SGC can be regarded as a fixed feature extraction/smoothing
component $\tilde{\mathbf{X}}=\mathbf{P}^{K}\mathbf{\mathbf{X}}$
followed by a linear logistic regression classifier $\tilde{\mathbf{Y}}=\mathrm{softmax}\left(\tilde{\mathbf{\mathbf{X}}}\mathbf{W}\right)$.
In \cite{GCN-simplifying-wu-ICML-2019}, Wu \emph{et al.} suggested
that SGC acts as a low-pass filter and such a signified model performs
comparably to many SOTA models.

\subsection{APPNP}

Personalized propagation of neural predictions (PPNP) and its fast
approximation, APPNP \cite{APPNP-klicpera-ICLR-2019}, consider the
relationship between GCN and PageRank to derive an improved propagation
scheme based on personalized PageRank. Let $\mathbf{H}^{(0)}=h_{\theta}\left(\mathbf{X}\right)$
be a multilayer perceptron (MLP) parameterized by $\theta$. Then
PPNP is defined as
\begin{equation}
\tilde{\mathbf{Y}}=\mathrm{softmax}\left(\alpha\left(\mathbf{I}-(1-\alpha)\mathbf{P}\right)^{-1}\mathbf{H}^{(0)}\right),
\end{equation}
where $\alpha\in(0,1]$ is the teleport (or restart) probability of
the topic-sensitive PageRank.

Similar to SGC, PPNP separates the neural network used for generating
predictions from the propagation scheme. While APPNP further approximates
topic-sensitive PageRank via power iteration, i.e.,
\begin{equation}
\begin{array}{c}
\mathbf{H}^{(i)}=\alpha\mathbf{H}^{(0)}+\left(1-\alpha\right)\mathbf{P}\mathbf{H}^{(i-1)}\\
\tilde{\mathbf{Y}}=\mathrm{softmax}\left(\alpha\mathbf{H}^{(0)}+\left(1-\alpha\right)\mathbf{P}\mathbf{H}^{(K-1)}\right)
\end{array}.
\end{equation}
One of the great advantages of PPNP and APPNP is that they decouple
feature transformation and propagation procedures of the vanilla GCN
without increasing the number of trainable parameters.

\subsection{Classic Linear Models}

We broadly divide classic linear models into supervised methods and
unsupervised methods. Similar to SGC, a typical supervised linear
model can be treated as a fully linearized MLP, given by 
\begin{equation}
\mathbf{\mathbf{X}}_{\mathcal{T}}\mathbf{W}_{1}\mathbf{W}_{2}\cdots\mathbf{W}_{\ell}=\mathbf{\mathbf{X}}_{\mathcal{T}}\mathbf{W}=\mathbf{Y}_{\mathcal{T}}.\label{eq:lin-semi}
\end{equation}
Here, $\mathbf{\mathbf{X}}_{\mathcal{T}}$ and $\mathbf{Y}_{\mathcal{T}}$
denote training samples and corresponding target matrix, respectively.
Such a model is also known as a ridge regression classifier \cite{K-RR-AN-CVPR-2007,Graph-RR-HSIAnalysis-HangRL-JSTARS-2017}.
Besides, logistic regression and softmax regression are its two most
frequently-used variants in deep neural networks \cite{HSIC-New-Frontier-Ghamisi-IGRSM-2018,GNN-Survey-ZHNAG-TKDE-2020,Deep-learning-LeCun-Nature-2015,CNN-overview-GuJX-PR-2018,DCELM-CAI2020-NEUCOM}.

The unsupervised fashion of a linear model often follows a common
assumption, i.e., data points lie in a union of linear subspaces.
While the subspace representation coefficients $\mathbf{Z}\in\mathbb{R}^{N\times N}$
can be obtained by solving the following linear self-expressive model,
i.e., 
\begin{equation}
\mathbf{\mathbf{X}}^{T}\mathbf{Z}=\mathbf{X}^{T},~s.t.~diag\left(\mathbf{Z}\right)=0\label{eq:lin-uns}
\end{equation}
Notably, the main difference between Eq. \eqref{eq:lin-semi} and
Eq. \eqref{eq:lin-uns} is that the former considers the combination
between every feature, while the latter considers samples. In order
to achieve an effective solution, various norm constraints are often
imposed on $\mathbf{W}$ or $\mathbf{Z}$. Sparse Subspace Clustering
(SSC) \cite{SSC-Elhamifar-TPAMI-2013} utilizes an $\ell_{1}$ norm
$\left\Vert \mathbf{Z}\right\Vert _{1}$, while Low Rank Subspace
Clustering (LRSC) \cite{LRSC-1-Vidal-PRLetters-2014} adopts a nuclear
norm $\left\Vert \mathbf{Z}\right\Vert _{*}$, just to name a few.
Despite their success, the objective functions derived from these
constraints are not smooth, leading to inefficient solutions. In contrast,
the Frobenius norm $\left\Vert \mathbf{Z}\right\Vert _{F}$ will result
in a closed-form solution for linear models.
\begin{figure*}[tbh]
\begin{centering}
\subfloat[Graph-form Illustration of our FLGC framework. FLGC first performs
parameter-free propagation which aggregates multi-hop neighboring
information with a power iteration of propagation matrix $\mathbf{P}$
and then predicts node labels by calculating a closed-form solution
(affinity matrix). ]{\begin{centering}
\includegraphics[width=1.8\columnwidth]{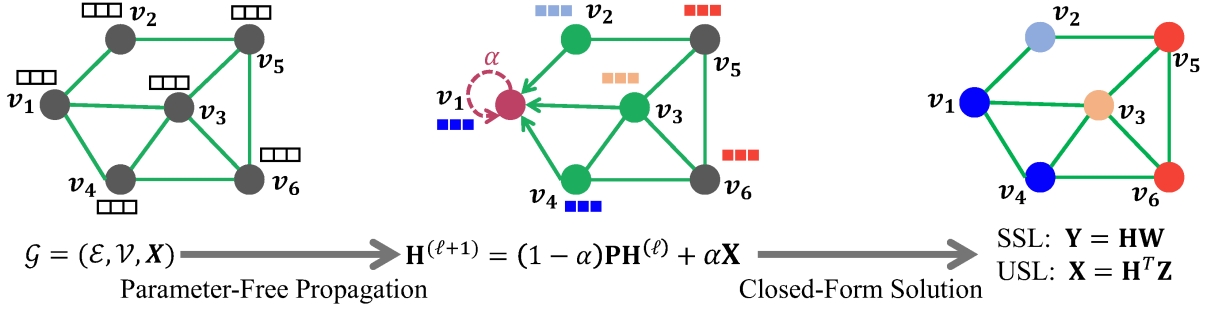}
\par\end{centering}
}
\par\end{centering}
\begin{centering}
\subfloat[Matrix-form data flow of FLGC. It can be seen that FLGC conducts a
personalized multi-hop propagation operation with a fixed propagation
matrix $\mathbf{P}$, followed by a linear model parameterized by
a weight matrix $\mathbf{W}$ for semi-supervised classification or
a coefficient matrix $\mathbf{Z}$ for unsupervised classification. ]{\begin{centering}
\includegraphics[width=1.8\columnwidth]{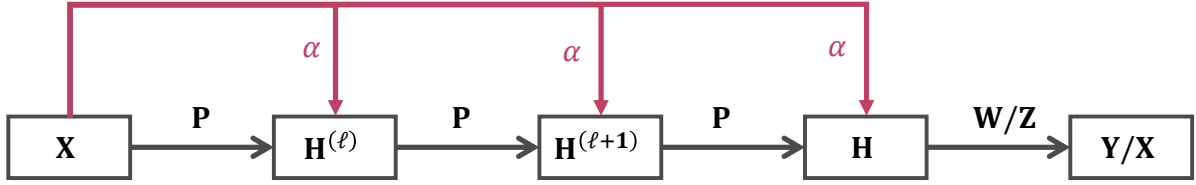}
\par\end{centering}
}
\par\end{centering}
\caption{Schematic representations of our proposed FLGC. \label{fig:Illustration}}
\end{figure*}

\section{Fully Linear Graph Convolutional Networks\label{sec:FLGC}}

We present the general FLGC framework for semi-supervised classification
and unsupervised classification problems, as illustrated in Fig. \ref{fig:Illustration}
(a). The core behind our FLGC is to generalize GCNs to traditional
linear models so that a) training linear GCN model with global optimal
solution b) enabling traditional linear models to work on graph-structured
data, 3) further simplifying the existing GCN models.

\subsection{Fully Linearization of GCN}

Inspired by the SGC \cite{GCN-simplifying-wu-ICML-2019} model, we
further remove all nonlinear operations of a $K$-layer GCN, including
the logistic regression classifier\footnote{It should be noticed that both Logistic (Sigmoid) and Softmax function
are often used as nonlinear activation in neural networks. Thus, we
consider SGC to be not fully linear. That is why it cannot calculate
a closed-form solution.}. It derives the following linear GCN
\begin{equation}
\mathbf{P}^{K}\mathbf{\mathbf{X}}\mathbf{W}=\mathbf{\tilde{Y}}.\label{eq:lin-gcn}
\end{equation}
This linearization brings three major benefits. First, it is easy
to decouple the fully linear GCN into two stages: a parameter-free
feature propagation stage (i.e., $\mathbf{H}=\mathbf{P}^{K}\mathbf{\mathbf{X}}$,
where $\mathbf{H}$ denotes graph embedding) and a target-dependent
prediction stage (i.e., $\mathbf{\tilde{Y}}=\mathbf{H}\mathbf{W}$).
The former aggregates $K$-hop neighborhoods based on a predefined
propagation matrix $\mathbf{P}$. While the later acts as a ridge
regression classifier parameterized by $\mathbf{W}$. Second, it establishes
a relationship between traditional linear models and GCN models. This
relationship enables us to reconsider the traditional linear models
from the graph representation learning point of view. Also, it endows
the classic ridge regression classifier with the ability to handle
graphs directly. Third, the linearization makes it possible to efficiently
solve the global optimal solution of GCN without using gradient descent.
We refer to such GCN as Fully Linear Graph Convolution (FLGC). A matrix-form
data flow of FLGC is depicted in Fig. \ref{fig:Illustration} (b).
More details will be introduced as follows.

\subsection{Multi-hop Propagation}

In light of the aforementioned linearization, we can define various
propagation strategies and incorporate them into Eq. \eqref{eq:lin-gcn}.
Here we introduce a stable propagation scheme for our FLGC. According
to the spectral graph theory, $\underset{K\rightarrow\infty}{\lim}\mathbf{P}^{K}\mathbf{X}$
will converge to a stationary state as the number of propagation steps
increases \cite{APPNP-klicpera-ICLR-2019,GCNII-Chen-ICML-2020}. While
the node representations on the same connected component of a graph
become indistinguishable, i.e., over-smoothing problem \cite{APPNP-klicpera-ICLR-2019,GCNII-Chen-ICML-2020}.
Moreover, $\mathbf{P}^{K}\mathbf{X}$ serves as a structure aggregation
term and ignores the importance of initial node features. The initial
node features often imply unique discriminant information, especially
for those data without directly available graph structures. 

To balance the structure between initial features, we adopt an extended
APPNP's propagation scheme to propagate multi-hop neighboring information.
The propagation procedure with $K$ power iteration is recursively
calculated by 
\begin{equation}
\begin{array}{c}
\mathbf{H}^{\left(\ell+1\right)}=\left(1-\alpha\right)\mathbf{P}\mathbf{H}^{\left(\ell\right)}+\alpha\mathbf{X}\\
s.t.~\mathbf{H}^{\left(1\right)}=\mathbf{X},~\ell=1,\cdots,K
\end{array}.\label{eq:appnp-1}
\end{equation}
Instead of using a neural network to generate a prediction as it is
done in APPNP \cite{APPNP-klicpera-ICLR-2019}, we directly use the
initial $\mathbf{X}$ as topics to be ranked in the topic-sensitive
PageRank \cite{PageRank-page-1999}. Let the resulting final graph
embedding be $\mathbf{H}=\mathbf{H}^{\left(K\right)}$. This ensures
that $\mathbf{H}$ is always contributed by both structure and initial
features with a fixed proportion $\alpha$. It is trivial to prove
that SGC's propagation (i.e., Eq. \eqref{eq:lin-gcn}) is a special
case of Eq. \eqref{eq:appnp-1} with $\alpha=0$. Furthermore, when
$\alpha=1$, SGC degrades into an ordinary neural network, in which
no structural information is used. From the residual network's point
of view \cite{GCN-kipf-ICLR-2017,GCNII-Chen-ICML-2020,ResNet-He_CVPR-2016},
our propagation scheme is a special form of residual connection, where
each forward step connects with initial inputs and weighted by $\alpha$
{[}as shown in Fig. \ref{fig:Illustration} (b){]}. Thus, our propagation
mechanism is also called initial residual \cite{GCNII-Chen-ICML-2020}.

\subsection{FLGC for Semi-Supervised Learning}

Having introduced the FLFC framework, we are ready to calculate the
closed-form solution for a specific downstream task. We first use
the FLGC for the semi-supervised node classification problem. Let
$\hat{\mathbf{Y}}\in\mathbb{R}^{N\times C}$ be an augmented target
matrix, where labeled nodes are presented as one-hot vectors while
unlabeled nodes are marked as zero vectors. Further let $\mathbf{M}$
be a diagonal mask matrix associated with the labeled and unlabeled
nodes. Thus, the semi-supervised FLGC can be denoted by 

\begin{equation}
\mathbf{H}\mathbf{W}=\mathbf{\hat{\mathbf{Y}}}.
\end{equation}
To effectively solve this problem, we rewrite the objective function
as a Frobenius norm minimization problem, i.e., 

\begin{equation}
\underset{\mathbf{W}}{\arg\min}\left\Vert \mathbf{W}\right\Vert _{F},~s.t.~\mathbf{H}\mathbf{W}=\widetilde{\mathbf{Y}}.
\end{equation}
The problem can be further expressed in the following form by using
the Lagrangian multiplier

\begin{equation}
\begin{aligned}\underset{\mathbf{W}}{\arg\min~}\mathcal{L}= & \frac{1}{2}\left\Vert \mathbf{M}^{1/2}\left(\mathbf{H}\mathbf{W}-\mathbf{\hat{\mathbf{Y}}}\right)\right\Vert _{F}^{2}+\frac{\lambda}{2}\left\Vert \mathbf{W}\right\Vert _{F}^{2}\\
= & \frac{1}{2}\mathrm{tr}(\mathbf{W}^{T}\mathbf{H}{}^{T}\mathbf{M}\mathbf{H}\mathbf{W}+\mathbf{\hat{\mathbf{Y}}}^{T}\mathbf{M}\mathbf{\hat{\mathbf{Y}}}-\\
 & 2\mathbf{\hat{\mathbf{Y}}}^{T}\mathbf{M}\mathbf{H}\mathbf{W}+\lambda\mathbf{W}^{T}\mathbf{W})
\end{aligned}
,
\end{equation}
where $\lambda$ denotes a regularization coefficient. The partial
derivative of $\mathcal{L}$ with respect to $\mathbf{W}$ is 

\begin{equation}
\begin{aligned}\frac{\partial\mathcal{L}}{\partial\mathbf{W}}= & \left(\mathbf{H}^{T}\mathbf{M}\mathbf{H}\mathbf{W}-\mathbf{H}^{T}\mathbf{M}\mathbf{\hat{\mathbf{Y}}}+\lambda\mathbf{W}\right)\end{aligned}
.
\end{equation}
The global optimal solution is derived when $\frac{\partial\mathcal{L}}{\partial\mathbf{W}}=0$,
i.e., 
\begin{equation}
\begin{aligned}\mathbf{H}^{T}\mathbf{M}\mathbf{H}\mathbf{W}-\mathbf{H}^{T}\mathbf{M}\mathbf{\hat{\mathbf{Y}}}+\lambda\mathbf{W} & =0\end{aligned}
\label{eq:tran-solu-1}
\end{equation}
Consequently, we denote the solution in closed form as
\begin{equation}
\begin{aligned}\boldsymbol{\mathbf{W}}^{*}=\left(\mathbf{H}^{T}\mathbf{M}\mathbf{H}+\lambda\mathbf{I}_{D}\right)^{-1}\mathbf{H}^{T}\mathbf{M}\mathbf{\hat{\mathbf{Y}}}.\end{aligned}
\label{eq:tran-solu}
\end{equation}
Finally, we infer the test node's labels via a transductive manner. 

\subsection{FLGC for Clustering}

Our unsupervised FLGC follows the classic subspace clustering by assuming
that the $K$-hop graph embeddings lie in a union of $C$ subspaces.
More precisely, every node belonging to a certain subspace can be
explicitly represented using other nodes in this subspace while subspaces
do not interact with each other. We refer to this property of data
as self-expressiveness. However, using an initial input to model such
a property may lead to an unreliable estimate of subspace coefficients
due to outliers and noisy points. Instead, we model our unsupervised
FLGC based on the graph embedding of inputs. The motivation behind
our method is that the intra-class variation of the initial inputs
can be smoothed by using the graph convolution. Formally, we express
our unsupervised FLGC as follows
\begin{equation}
\underset{\mathbf{Z}}{\arg\min~}\left\Vert \mathbf{Z}\right\Vert _{q},~s.t.~\mathbf{H}^{T}\mathbf{Z}=\mathbf{X}^{T},\mathrm{diag}\left(\mathbf{Z}\right)=0
\end{equation}
Here, $\mathbf{Z}\in\mathbb{R}^{N\times N}$ is an affinity matrix,
in which the $j$-th column denotes the representation coefficient
vector of node $v_{j}$, and $q$ indicates the $q$-norm of a matrix.
By analogy with classic subspace clustering models, $q=1$ will lead
to a standard sparse subspace representation while replacing $q$
with the nuclear norm will derive a low-rank subspace representation. 

In this paper, we aim to calculate a dense subspace representation
by adopting the Frobenius norm, as well as maintaining the consistency
of our FLGC framework. It has been proven \cite{EDSC-PanJ-WACV-2014}
that the constraint $\mathrm{diag}\left(\mathbf{Z}\right)=0$ can
be relaxed and discarded by using a Frobenius norm. Hence, our unsupervised
FLGC can be rewritten as
\begin{equation}
\begin{aligned}\underset{\mathbf{Z}}{\arg\min~}\mathcal{L}=\frac{1}{2}\left\Vert \mathbf{H}^{T}\mathbf{Z}-\mathbf{X}^{T}\right\Vert _{F}^{2}+\frac{\lambda}{2}\left\Vert \mathbf{Z}\right\Vert _{F}^{2}=\\
\mathrm{tr}\left(\mathbf{Z}^{T}\mathbf{H}\mathbf{H}^{T}+\mathbf{X}^{T}\mathbf{X}-2\mathbf{X}\mathbf{H}^{T}\mathbf{Z}+\frac{\lambda}{2}\mathbf{Z}^{T}\mathbf{Z}\right)
\end{aligned}
.\label{eq:un-flgc}
\end{equation}
We further compute $\frac{\partial\mathcal{L}}{\partial\mathbf{W}}$
as 
\begin{equation}
\frac{\partial\mathcal{L}}{\partial\mathbf{Z}}=\left(\mathbf{H}\mathbf{H}^{T}\mathbf{Z}-\mathbf{H}\mathbf{X}^{T}+\lambda\mathbf{Z}\right).
\end{equation}
Similar to the semi-supervised FLGC, we can give the global optimal
closed-form solution of Eq. \eqref{eq:un-flgc}, i.e., 
\begin{equation}
\boldsymbol{\mathbf{Z}}^{*}=\left(\mathbf{H}\mathbf{H}^{T}+\lambda\mathbf{I}_{N}\right)^{-1}\mathbf{H}\mathbf{X}^{T}.\label{eq:clu-solu}
\end{equation}
Following \cite{HSI-Clustering-GCSC-CAI-TGRS-2020} and \cite{HSI-Clustering-GCSC-CAI-TGRS-2020},
we perform the spectral clustering on $\boldsymbol{\mathbf{Z}}^{*}$
to segment subspaces after using a block-structure heuristic. 

\subsection{Remarks on FLGC}

In Algorithm ref{alg:pesudocode}, we provide the pseudocode for our
semi-supervised and unsupervised FLGC. Noticed that both methods share
a unified learning procedure and are easy to implement and train.
In reality, our proposed FLGC models can be treated as natural generalizations
of classic linear models in the non-Euclidean domain. \begin{table}[!htbp]  
\centering 
\label{alg:pesudocode}	 
\begin{tabular}{l}  
\toprule  
\textbf{Algorithm 1} Pseudocode of FLGC in a PyTorch-like style.\\ 
\midrule 
\\ 
\texttt{\textbf{\textcolor{teal}{\# lambda: regularization coefficient}}}\\ 
\texttt{\textbf{\textcolor{teal}{\# alpha: teleport (or restart) probability}}}\\ 
\texttt{\textbf{\textcolor{teal}{\# }}}\\
\texttt{\textbf{\textcolor{teal}{\# fit: calculate closed form solutions}}}\\ 
\texttt{\textbf{\textcolor{teal}{\# sc: spectral clustering}}}\\
\texttt{\textbf{\textcolor{teal}{\# mm: matrix-matrix multiplication}}}\\
\texttt{\textbf{\textcolor{teal}{\# gcn\_norm: normlize adjacency matrix}}}\\ 
\\
\texttt{\textbf{\textcolor{teal}{\# compute augmented normalized adjacency}}}\\
\texttt{\textbf{P = gcn\_norm(A)}} \texttt{\textbf{\textcolor{teal}{ \# NxD}}}\\
\\
\texttt{\textbf{\textcolor{teal}{\# compute K-hop graph embedding}}}\\
\texttt{\textbf{H = X}} \texttt{\textbf{\textcolor{teal}{ \# NxD}}}\\
\texttt{\textbf{for i range(K):}}\\
\texttt{\textbf{\textcolor{teal}{\hspace{1.5em}\# propagate neighborhood using Eq.(10)}}}\\ 
\texttt{\textbf{\hspace{1.5em}H = (1 - alpha) * mm(A\_hat, H) + alpha * X}} \texttt{\textbf{\textcolor{teal}{ \# NxD}}}\\
\\
\texttt{\textbf{\textcolor{teal}{\# SEMI-SUPERVISED FLGC}}}\\
\texttt{\textbf{if task == 'ssl':}} \\ 
\texttt{\textbf{\textcolor{teal}{\hspace{1.5em}\# compute W using Eq.(16) }}}\\ 
\texttt{\textbf{\hspace{1.5em}W = fit(H, Y\_aug, M)}} \texttt{\textbf{\textcolor{teal}{ \# DxC}}}\\
\texttt{\textbf{\textcolor{teal}{\hspace{1.5em}\# predict unlabeled nodes}}}\\ 
\texttt{\textbf{\hspace{1.5em}y = mm(H, W)}} \\
\\ 
\texttt{\textbf{\textcolor{teal}{\# UNSUPERVISED FLGC}}}\\
\texttt{\textbf{elif task == 'usl':}}\\
\texttt{\textbf{\textcolor{teal}{\hspace{1.5em}\# compute Z using Eq.(20) }}}\\ 
\texttt{\textbf{\hspace{1.5em}Z = fit(H, X)}}\\
\texttt{\textbf{\textcolor{teal}{\hspace{1.5em}\# assign node labels through spectral clustering}}}\\ 
\texttt{\textbf{\hspace{1.5em}y = sc(Z)}}\\ 
\\
\bottomrule 
\end{tabular}  
\end{table}

Let $\mathcal{F_{\varTheta}}\left(\mathcal{G}\right)$ be the generalized
linear learning model defined on a graph $\mathcal{G}$. FLGC can
be broadly denoted as a classic model, $f_{\varTheta}$, multiplied
by a $K$-hop propagation matrix, $\mathcal{P}_{K}$, i.e., 
\begin{equation}
\mathcal{F_{\varTheta}}\left(\mathcal{G}\right)=\mathcal{P}_{K}\left(\mathbf{A}\right)f_{\varTheta}\left(\mathbf{X}\right).
\end{equation}
The only trainable parameter, $\varTheta$, is derived from the classic
model which can be solved in a similar manner to the existing models.
It is easy to prove that FLGC is a generalization of classic linear
models in the non-Euclidean domain. When $\mathcal{P}_{K}\left(\mathbf{A}\right)=\mathbf{I}$,
signifying that $\mathcal{G}$ does not have any edges except for
self-loops, FLGC is equivalent to classic linear models. Benefited
from the decoupled design, $\mathcal{P}_{K}\left(\mathbf{A}\right)$
is target-independent and can be precomputed. Thus, FLGC does not
increase the computational burden compared to the classic linear models.
Here, we show the connections of FLGC to existing models. 
\begin{itemize}
\item \textbf{FLGC v.s. Manifold Regularization} Let $\mathcal{L}_{MR}=\mathcal{O}_{emp}\left(f\right)+\lambda\mathcal{R}_{srm}\left(\varTheta\right)+\gamma\mathcal{R}_{pri}\left(f\right)$
be the objective function of a manifold regularized model, where $\mathcal{O}_{emp}$,
$\mathcal{R}_{srm}$, and $\mathcal{R}_{pri}$ denote the empirical
error term, the structural risk term, and the manifold prior term,
orderly. Further let $\mathcal{L}_{FLGC}=\mathcal{O}_{emp}\left(\mathcal{F}\right)+\lambda\mathcal{R}_{srm}\left(\varTheta\right)$
be the proposed FLGC. As suggested in \cite{UnifyingGCN-XiaoWang-WWW-21},
both $\mathcal{L}_{MR}$ and $\mathcal{L}_{FLGC}$ are derived from
the same optimization framework. Nonetheless, there is a considerable
difference between them. That is, our FLGC directly works in the non-Euclidean
domain, while the manifold regularization is proposed for Euclidean
data. During the learning, manifold regularized models use the graph
structure as the prior knowledge, generally defined as $\mathcal{R}_{pri}=\mathrm{tr}\left(f^{T}\mathbf{L}f\right)$,
which is incorporated as a single term balanced by an additional regularization
coefficient. In particular, the manifold regularization cannot model
the long-range relationships. In contrast, our FLGC propagates multi-hop
structural  information in a more general and flexible manner. 
\item \textbf{FLGC v.s. AutoEncoder} Our unsupervised FLGC is highly related
to linear autoencoders \cite{HSI-Clustering-GCSC-CAI-TGRS-2020}.
We define a linear graph autoencoder as $\mathcal{D}_{\mathbf{Z}}\left(\mathcal{E}_{\varTheta}\left(\mathcal{G}\right)\right)=\mathbf{X}$,
where $\mathcal{E}_{\varTheta}$ and $\mathcal{D}_{\mathbf{Z}}$ are
the decoder and encoder, respectively. By collapsing weights matrices
of a $K$-layer encoder in a single matrix $\varTheta$, an autoencoder
with a self-expressive layer becomes $\mathcal{E}_{\varTheta}=\varTheta\mathbf{X}^{T}\mathbf{P}^{K}=\mathbf{H}$
and $\mathcal{D}_{\mathbf{Z}}=\mathbf{H}\mathbf{Z}=\mathbf{X}$. By
replacing $\varTheta$ using a fixed unitary matrix $\mathbf{I}$,
such encoder becomes the propagation stage of the unsupervised FLGC,
while our $\mathcal{D}_{\mathbf{Z}}$ is a single layer self-expression.
Furthermore, our FLGC aims to reconstruct node features rather than
the structure that adopted in the graph autoencoder \cite{GAE-Kipf-2016}.
\end{itemize}

\section{Experiments of SSL \label{sec:Experiments-of-Traductive}}

In this section, we compare the proposed semi-supervised FLGC model
against many SOTAs and classic methods on 3 real-world graph benchmark
datasets and 24 regular UCI datasets. Furthermore, numerous ablation
experiments are conducted to systematically and comprehensively analyze
the effectiveness and robustness of the proposed FLGC.\begin{table}[htbp]   
\centering   
\caption{Summary of citation network datasets.} 
\label{tab:citation-dataset}  
\begin{tabular}{l|ccccc}     
\toprule     
Dataset &  \#Nodes  & \#Edges  &\#Classes & \#Features & Train/Dev/Test \\     
\midrule     
Cora  & 2,708 & 5,429  & 7 & 1433 &  140/500/1,000 \\     
Citeseer & 3,327 & 4,732  & 6 & 3703 &  120/500/1,000 \\     
Pubmed & 19,717 & 44,338  & 3 & 500 &  60/500/1,000 \\     
\bottomrule     
\end{tabular}  
\end{table}\begin{table}[!htbp] 
\centering 
\caption{Summary of baseline comparison under public splits for Cora, Citeseer, and Pubmed. FLGC* corresponds to FLGC with $\alpha=0.1$. The results at the top are collected from literature while the middle is reproduced according to our settings.}
\label{tab:res-public-graph}	
\begin{tabular}{rccc} 
\toprule       
Method & Cora & Citeseer & PubMed  \\ 
\midrule GCN & 81.4 $\pm$0.4 & 70.9 $\pm$0.5 & 79.0 $\pm$0.4 \\ 
GAT & 83.3 $\pm$0.7 & 72.6 $\pm$0.6 & 78.5 $\pm$0.3 \\ 
FastGCN  & 79.8 $\pm$0.3 & 68.8 $\pm$0.6  & 77.4 $\pm$0.3 \\ 
GIN & 77.6 $\pm$1.1 & 66.1 $\pm$0.9 & 77.0 $\pm$1.2 \\ 
LNet & 80.2 $\pm$3.0 & 67.3 $\pm$0.5 & 78.3 $\pm$0.6 \\ 
AdaLNet & 81.9 $\pm$1.9 & 70.6 $\pm$0.8 & 77.8 $\pm$0.7\\ 
DGI& 82.5 $\pm$0.7 & 71.6 $\pm$0.7 & 78.4 $\pm$0.7 \\ 
SGC & 81.0$\pm$0.0 & 71.9 $\pm$0.1 & 78.9 $\pm$0.0 \\ 
MixHop & 81.9$\pm$ 0.4 & 71.4$\pm$0.8 & 80.8$\pm$0.6\\ 
DropEdge & 82.8& 72.3 & 79.6 \\ 
G$^3$NN & 82.5$\pm$0.2 & 74.4$\pm$0.3 & 77.9 $\pm$0.4\\
\midrule
GCN & 81.1$\pm$0.2 & 69.8$\pm$0.2 & 79.4$\pm$0.1\\
GCN-Linear & 80.8$\pm$0.0 & 68.7$\pm$0.1 & 79.4$\pm$0.1 \\
SGC & 81.7$\pm$0.0 & 71.1$\pm$0.0 & 76.6$\pm$0.0 \\
APPNP & 82.5$\pm$0.2 & 70.4$\pm$0.1 & 79.4$\pm$0.3 \\
\midrule
FLGC & 82.9$\pm$0.0 & 72.7$\pm$0.0 & 79.2$\pm$0.0 \\
FLGC* & \textbf{84.0$\pm$0.0} & \textbf{73.2$\pm$0.0} & \textbf{81.1$\pm$0.0} \\
\bottomrule
\end{tabular} 
\end{table}\begin{table*}[!htbp] 	
\centering 	
\caption{Summary of baseline comparison  under random splits and full splits for Cora, Citeseer, and Pubmed.}
\label{tab:res-random-graph}		
\begin{tabular}{rcccccc} 		
\toprule 
\multirow{2}{*}{Method}
& \multicolumn{3}{c}{Random Split} & \multicolumn{3}{c}{Full Split} \\ 
\cmidrule(r){2-4}   \cmidrule(r){5-7} 
& Cora  & Citeseer & PubMed & Cora  & Citeseer & PubMed \\ 
\midrule 
GCN   & 79.1$\pm$1.8 & 67.9$\pm$1.2 & 76.9$\pm$2.9 & 86.4  & 75.4  & 85.9 \\ 		
GCN-Linear & 79.8$\pm$2.1 & 68.4$\pm$2.1 & 76.7$\pm$1.4 & 87.0    & 75.7  & 83.9 \\ 		
SGC   & 81.3$\pm$1.7 & 68.5$\pm$2.2 & 76.4$\pm$3.3 & 86.2  & 77.6  & 83.7 \\ 		
APPNP & 81.0$\pm$0.0 & 68.5$\pm$0.0 & 75.1$\pm$0.0 & 88.4  & 78.6  & 82.3 \\ 		
FLGC  & 81.5$\pm$0.6 & 71.0$\pm$0.9 & 77.6$\pm$0.3 & 87.0    & 78.1  & 87.9 \\ 		
FLGC* & \textbf{82.0$\pm$0.1} & \textbf{72.1$\pm$0.0} & \textbf{77.7$\pm$0.0} & \textbf{88.5}  & \textbf{79.2}  & \textbf{88.3} \\ 
\bottomrule
\end{tabular} 	
\end{table*}
\begin{figure*}[tbh]
\begin{centering}
\subfloat[]{\includegraphics[width=0.66\columnwidth]{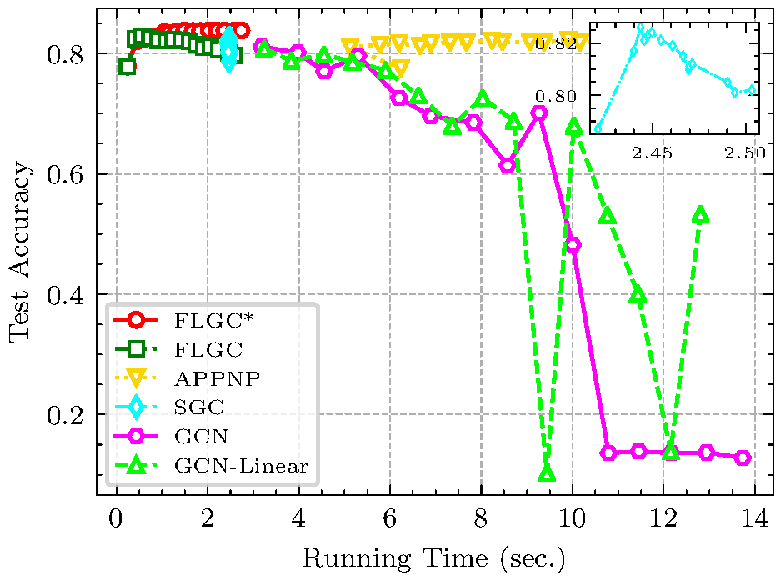}}\subfloat[]{\includegraphics[width=0.66\columnwidth]{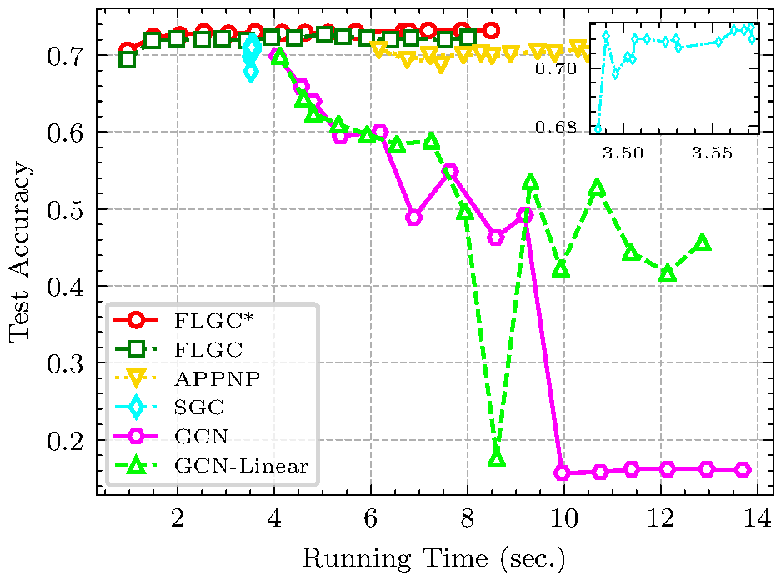}}\subfloat[]{\includegraphics[width=0.66\columnwidth]{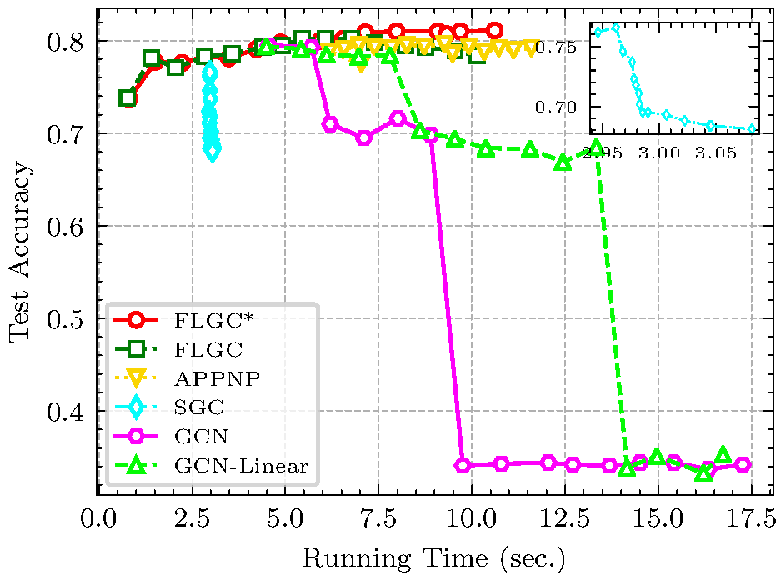}}
\par\end{centering}
\caption{Classification accuracy vs. training time with varying propagation
step on (a) Cora, (b) Citeseer, and (c) Pubmed datasets. We keep other
hyperparameters fixed except for $K$ for each model so that running
time is affected only by the propagation step. In the figures, each
curve contains 15 points (i.e., 15 propagation steps) and each corresponds
to the running time (x-axis) and test accuracy (y-axis) yielded by
a specific $K$-hop model, where $K=\left[1,2,\cdots,15\right]$ and
increases with running time. The\textbf{ }insets indicate the zoom-up
of SGC. \label{fig:graph-K-time}}
\end{figure*}
\begin{figure*}[tbh]
\begin{centering}
\subfloat[]{\includegraphics[width=0.66\columnwidth]{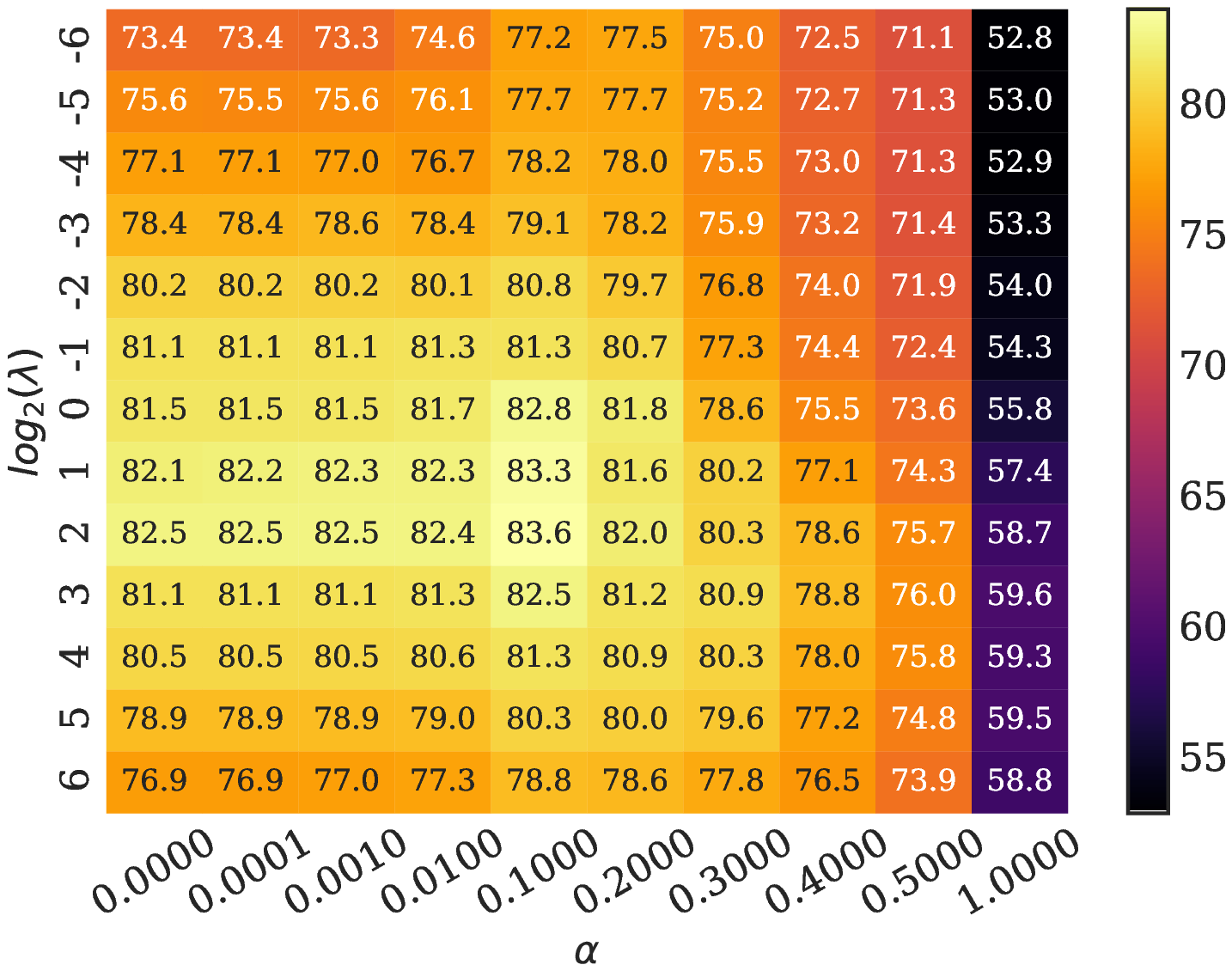}}\subfloat[]{\includegraphics[width=0.66\columnwidth]{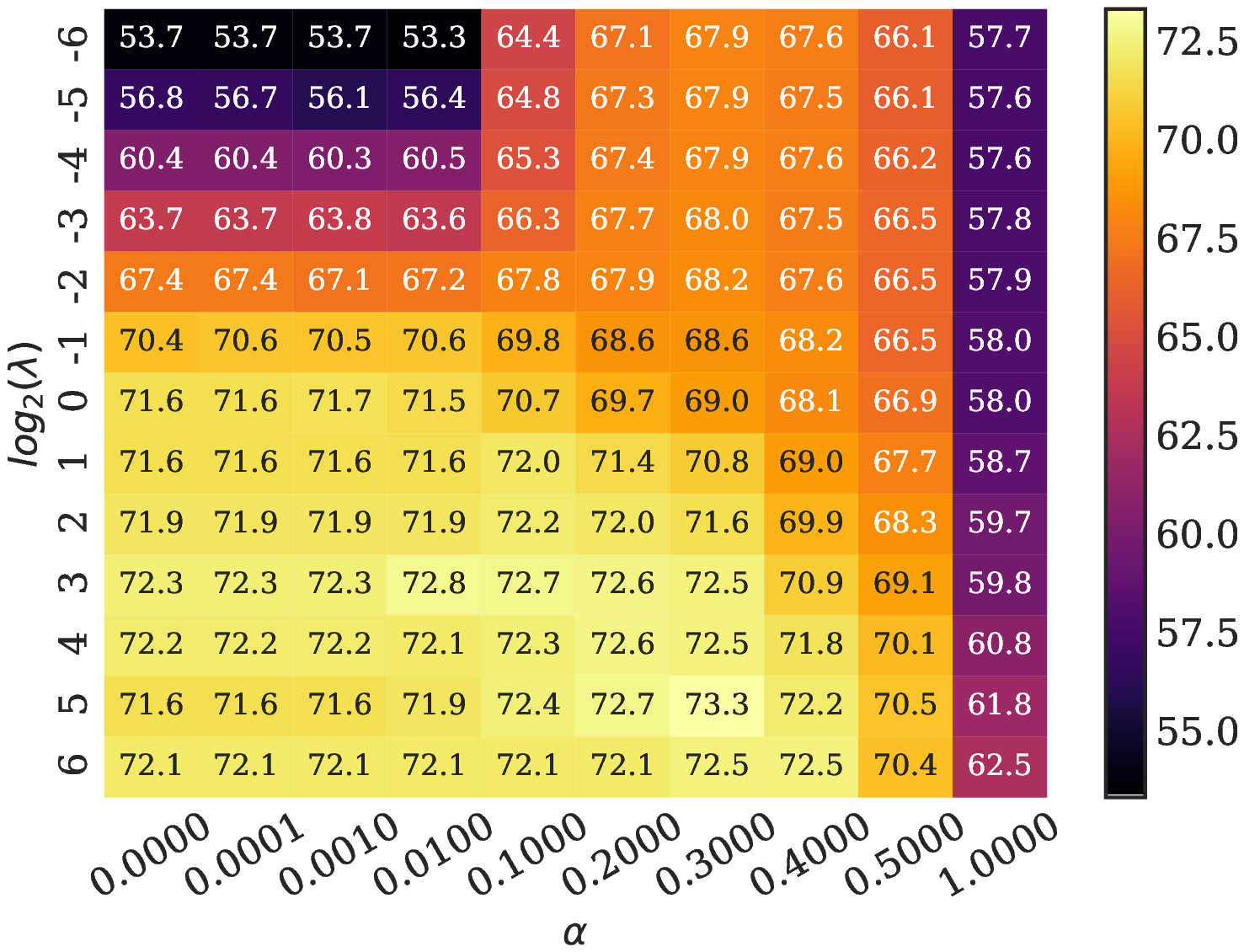}}\subfloat[]{\includegraphics[width=0.66\columnwidth]{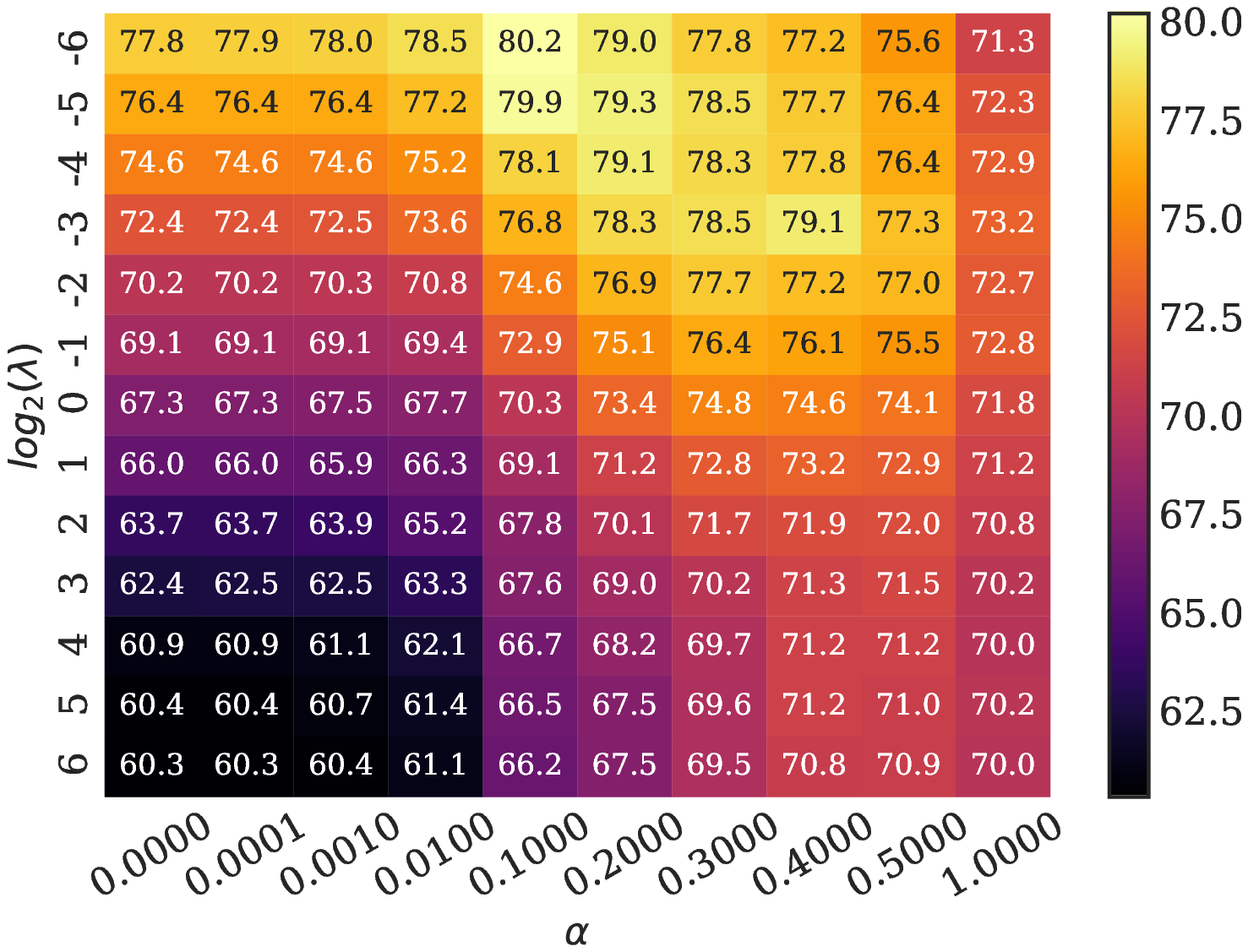}}
\par\end{centering}
\caption{Effect of $\lambda$ and $\alpha$ on (a) Cora, (b) Citeseer, and
(c) Pubmed datasets. \label{fig:graph-alpha-lapbda}}
\end{figure*}

\subsection{Results on Real-World Benchmarks}

\subsubsection{Dataset Description}

We evaluate our proposed FLGCs on three standard citation network
datasets available from the PyTorch Geometric library, including the
Cora, Citeseer, and PubMed \cite{graph-data-Sen-AI-Maazine-2008}.
The summary of these datasets is reported in Table \ref{tab:citation-dataset}.
In these datasets, nodes correspond to documents, and edges correspond
to citations; each node feature corresponds to the bag-of-words representation
of the document and belongs to one of the academic topics \cite{graph-data-Yang-ICML-2016}.
Given a portion of nodes and their labeled categories, e.g., history
and science, the task is to predict the category for other unlabeled
nodes over the same graph. 

\subsubsection{Baselines and Setup}

For citation network datasets, the proposed FLGCs compare against
numerous SOTA graph neural network models, including the vanilla GCN
\cite{GCN-kipf-ICLR-2017}, GAT \cite{GAT-PETAR-ICLR-2018}, FastGCN
\cite{FastGCN-Chen-ICLR-2018}, GIN \cite{GIN-XU-ICLR-2018}, LNET,
AdaLNet \cite{LNet-AdaLNet-ICLR-2019}, DGI \cite{DGI-ICLR-Petar-2019},
SGC \cite{GCN-simplifying-wu-ICML-2019}, MixHop \cite{MxiHop-ICML-2019},
DropEdge \cite{DropEdge-ICLR-2019}, and G$^{3}$NN \cite{G3NN-Ma-NIPS-2019}.
For these models, we give their results reported in the corresponding
literature. Moreover, we reproduce the vanilla GCN w/o non-linear
activation (GCN or GCN-Linear), SGC, and APPNP \cite{APPNP-klicpera-ICLR-2019}.
In our reproduction, we follow the settings suggested in the corresponding
papers. Specifically, we implement GCN and GCN-Linear using two-layer
graph convolution each with $16$ hidden neurons, and apply an $L_{2}$
regularization with $\lambda=0.0005$ on trainable parameters. For
APPNP, we adopt a two-layer MLP, each of which contains $64$ hidden
neurons and $\lambda=0.005$. For a fair comparison, we discard other
training tricks involved in backpropagation except for weight decay. 

We implement two variants of our FLGC model with the PyTorch library\footnote{Relies on Pytorch Geometric 1.6.3.},
i.e., FLGC{*} indicates our method that uses our propagation mechanism
and FLGC denotes our model that uses the SGC propagation. The hyper-parameters
in our models are determined by a grid search among $\lambda=[2^{-8},2^{8}]$,
$\alpha=[0,1]$, and $K=[0,20]$. We train and test all baselines
with the same data splits and random seeds on an NVIDIA GTX 1080 Ti
GPU, and report the average accuracy over $20$ runs. In our experiment,
we provide three types of data splits, i.e., public splits as described
in \cite{graph-data-Sen-AI-Maazine-2008}, random splits where training/validation/test
sets are generated randomly with the same proportion as the public
splits, and full splits where all remaining nodes are considered as
the training set. 

\subsubsection{Comparison with SOTAs}

Table \ref{tab:res-public-graph} reports the classification accuracies
of node classification with public splits. The results shown in the
top part of Table \ref{tab:res-public-graph} are collected from \cite{GCN-kipf-ICLR-2017,GAT-PETAR-ICLR-2018,FastGCN-Chen-ICLR-2018,GIN-XU-ICLR-2018,LNet-AdaLNet-ICLR-2019,DGI-ICLR-Petar-2019,GCN-simplifying-wu-ICML-2019,MxiHop-ICML-2019,DropEdge-ICLR-2019,G3NN-Ma-NIPS-2019}
while the results shown in the middle part of Table \ref{tab:res-public-graph}
are reproduced in our experiment. It can be seen that our FLGC models
consistently achieve large-margin outperformance across all datasets.
Through a point-by-point comparison, FLGC improves upon SGC by a margin
of $1.2\%$, $1.6\%$, and $2.6\%$ (absolute differences) on Cora,
Citeseer, and Pubmed, respectively, while the margins improved by
FLGC{*} upon APPNP are $1.5\%$, $2.8\%$, and $1.7\%$, respectively.
Through a vertical comparison, FLGC{*} achieves $1.1\%$, $0.5\%$
and $1.9\%$ improvement over FLGC, respectively. 

In Table \ref{tab:res-random-graph}, we further report the comparison
results using the random splits and full splits. We can observe that
the proposed FLGCs collectively outperform the competitors in terms
of average classification accuracy. It should be noted that our FLGCs
tend to obtain a more stable result than other baselines because of
their ability to offer closed-form solutions. In a nutshell, the above
experiments demonstrate that our FLGC framework is capable of achieving
the SOTA performance. 

\subsubsection{Running Time and Over-smoothing Analysis}

Fig. \ref{fig:graph-K-time} depicts the interaction between training
time and classification accuracy. To ensure a fair comparison, all
the training times are collected from the same hardware. In particular,
the precomputation time of $\mathbf{H}$ is taken into account for
SGC and FLGCs. We use fixed hyperparameters to train each model and
let $K$ be the only variable increasing from $1$ to $15$. Thus,
the running time of each model will definitely increase with $K$,
and then will indirectly affect the test accuracy. Remarkably, GCN
and linear GCN suffer from unstable performance when $K\geq2$ because
of over-smoothing. On the contrary, both FLGC and FLGC{*} are robust
to the propagation steps. Especially, there is no significant over-smoothing
effect in FLGC{*} across three figures, signifying the effectiveness
of our propagation scheme. Also, it suggests that the residual connection
is helpful to address the over-smoothing problem.

For a given graph, the training time of a graph model is positively
associated with the propagation steps (or layers). Nonetheless, our
FLGCs show a notable advantage over other methods (e.g., GCN and APPNP).
Especially, it is almost no increase in the computation time using
our multi-hop propagation scheme. Instead, such a strategy improves
FLGC significantly, particularly with large $K$. Despite computation
efficiency, SGC suffers from over-smoothing and its training relies
on the optimizer and its parameter settings. In summary, our FLGC
model achieves a good balance between classification accuracy and
training time. \begin{table*}[htbp]    	  
\centering    	   
\caption{Details of 24 UCI benchmark datasets.}
\label{tab:dataset-uci} 	   
\begin{tabular}{lccccclccccc}      			  	 \toprule      			  	 Dataset & \#classes & \#instances & \#features & \#train & \#test & Dataset & \#classes & \#instances & \#features & \#train & \#test \\      			  	 \midrule air & 3 & 359 & 64 & 37 &  322 & appendicitis & 2     & 106   & 7     & 12    & 94 \\  ecoli & 8     & 336   & 7     & 38    & 298 & heart & 2     & 270   & 13    & 27    & 243 \\      	 iris  & 3     & 150   & 4     & 15    & 135 & cleve & 2     & 296   & 13    & 30    & 266 \\ 	 fertility & 2     & 100   & 9     & 11    & 89 & segmentation & 7     & 210   & 18    & 21    & 189 \\     	 wine  & 3     & 178   & 13    & 19    & 159 & x8d5k & 5     & 1000  & 8     & 100   & 900 \\        	 wdbc  & 2     & 569   & 30    & 58    & 511 & vote  & 2     & 435   & 16    & 44    & 391 \\     	 haberman & 3     & 306   & 3     & 32    & 274 & wbc   & 2     & 683   & 9     & 69    & 614 \\  	 spectf & 2     & 267   & 44    & 28    & 239 & WBC & 2     & 683  & 9    & 69   & 614 \\    cotton & 6     & 356   & 21    & 37    & 319 & breast & 2     & 277   & 9     & 29    & 248 \\     	 seeds & 3     & 210   & 7     & 21    & 189 & australian & 2     & 690   & 14    & 70    & 620 \\ 	 glass & 6     & 214   & 10    & 23    & 191 & diabetes & 2     & 768   & 8     & 78    & 690 \\     	 zoo    & 7     & 101   & 16    & 13    & 88 & dnatest & 3     & 1186  & 180   & 120   & 1066 \\    \bottomrule     	  	 \label{tab:datasets} 
\end{tabular}  
\end{table*}

\subsubsection{Impact of $\lambda$ and $\alpha$}

We study the sensitivity of $\lambda$ and $\alpha$ of FLGC. As depicted
in Fig. \ref{fig:graph-alpha-lapbda} (a)-(f), both $\lambda$ and
$\alpha$ have a significant effect on the accuracy. Due to the difference
in the neighborhood structure, the optimum of $\lambda$ and $\alpha$
will be varied for different datasets. Usually, a large $\lambda$
value tends to bring a compact model, while a small value increases
the risk of over-fitting. For Cora and Citeseer datasets, a large
$\lambda$ is desired by the FLGC models, while this value should
be small on the Pubmed dataset. It should be noticed that FLGC{*}
is equivalent to FLGC and the classic ridge regression classifier
when $\alpha=0$ and $\alpha=1$, respectively. It can be seen that
the best settings for this hyper-parameter is around $\alpha\in\left[0.01,0.2\right]$.
The teleport probability $\alpha$ serves as a proportion showing
to what extent original features contribute to the propagation. Our
further analysis revealed the following tendencies: First, compared
to the two endpoints ($\alpha=0$ and $\alpha=1$) in Fig. \ref{fig:graph-alpha-lapbda},
FLGC improves the classic ridge regression with a significant margin.
This means that structure information is pretty useful for the traditional
linear model. Second, the original node features are helpful to improve
FLGC, which makes it possible to aggregate higher hop neighbors.

\subsection{Results on Regular Datasets}

\subsubsection{Dataset Description}

To further explore the generalization ability of FLGCs, we conduct
a series of experiences on $24$ widely-used regular classification
datasets taken from the University of California at Irvine (UCI) repository\footnote{http://archive.ics.uci.edu/ml/index.php}.
These datasets include a number of binary-class and multi-class classification
problems. In the preprocessing, all datasets are scaled into the range
of $[0,1]$ using the min-max standardization technique. For each
dataset, we randomly take $10\%$ samples from each class as the labeled
set and the rest $90\%$ as the unlabeled set. A detailed description
of these datasets is provided in Table \ref{tab:dataset-uci}.

\begin{table*}[htbp]    
\centering    
\caption{Semi-supervised classification accuracy on 24 UCI datasets. Best results are in bold.} 
\label{tab:acc-res-uci}
\resizebox{2\columnwidth}{!}{    
\begin{tabular}{rccccccccccc}       			
\toprule       			
Dataset & SS-ELM & TSVM  & LapRLS & GCN-Linear & GCN   & DropEdge & SGC   & GCNII & APPNP & FLGC & FLGC*\\ 		
\midrule       			
air   & 76.23$\pm$4.08 & 80.19$\pm$3.65 & 76.09$\pm$6.08 & 90.09$\pm$4.09 & 90.12$\pm$3.80 & 86.43$\pm$4.13 & 78.94$\pm$3.71 & 89.16$\pm$2.53 & 90.31$\pm$4.17 & 90.65$\pm$2.89 & \textbf{91.02$\pm$3.44}\\      ecoli & 77.41$\pm$5.31 & 79.43$\pm$4.34 & \textbf{85.86$\pm$1.99} & 83.41$\pm$3.90 & 83.55$\pm$3.87 & 80.71$\pm$2.85 & 77.13$\pm$1.47 & 83.28$\pm$2.61 & 81.32$\pm$3.96 & 84.09$\pm$2.10 &  84.73$\pm$1.92\\       			iris  & 80.19$\pm$4.26 & 92.56$\pm$3.67 & 91.93$\pm$3.34 & 91.63$\pm$4.63 & 90.81$\pm$4.89 & 88.81$\pm$5.95 & 88.59$\pm$6.29 & 92.67$\pm$2.05 & 91.41$\pm$3.49 & 96.30$\pm$0.66  & \textbf{96.81$\pm$0.47}\\       			Fertility & 75.28$\pm$10.03 & 71.74$\pm$8.72 & 77.25$\pm$5.80 & 73.03$\pm$9.91 & 74.38$\pm$10.31 & 83.48$\pm$4.69 & \textbf{88.76$\pm$0.00} & 76.18$\pm$8.92 & 77.87$\pm$9.36 & \textbf{88.76$\pm$0.00}  & \textbf{88.76$\pm$0.00}\\       			wine  & 79.53$\pm$7.66 & 94.09$\pm$2.10 & \textbf{95.13$\pm$2.12} & 93.08$\pm$1.57 & 93.14$\pm$1.14 & 90.44$\pm$2.00 & 91.13$\pm$4.50 & 93.33$\pm$2.69 & 91.64$\pm$2.22 & 94.03$\pm$1.69  & 94.28$\pm$1.47\\       			wdbc  &  89.17$\pm$5.90 & 93.64$\pm$1.83 & 85.92$\pm$2.41 & 93.56$\pm$1.75 & 93.60$\pm$2.05 & 93.68$\pm$1.60 & 91.78$\pm$1.68 & 93.52$\pm$0.98 & 93.62$\pm$1.86 & 94.58$\pm$0.90  &  \textbf{95.60$\pm$0.67}\\       			Haberman & 71.73$\pm$2.44 & 62.57$\pm$5.73 & 70.02$\pm$3.15 & 70.51$\pm$2.15 & 70.40$\pm$2.83 & 67.96$\pm$4.15 & 73.72$\pm$0.00 & 70.88$\pm$2.05 & 69.71$\pm$2.63 & \textbf{73.72$\pm$0.00}  & \textbf{73.72$\pm$0.00}\\       			SPECTF & 77.41$\pm$1.83 & 74.33$\pm$3.41 & 49.77$\pm$4.59 & 77.24$\pm$2.55 & 78.49$\pm$1.93 & 76.99$\pm$4.24 & \textbf{79.50$\pm$0.00} & \textbf{79.50$\pm$0.00} & 78.58$\pm$2.07 & \textbf{79.50$\pm$0.00}  & \textbf{79.50$\pm$0.00}\\       			CAR   & 74.18$\pm$3.00 & \textbf{86.46$\pm$2.19} &  83.78$\pm$1.20 & 79.36$\pm$1.18 & 85.52$\pm$1.23 & 82.77$\pm$1.80 & 73.89$\pm$0.73 & 85.40$\pm$1.10 & 85.03$\pm$1.17 & 78.91$\pm$0.78  & 78.96$\pm$0.84\\       			cotton & 60.64$\pm$4.00 & 78.10$\pm$3.24 & 76.10$\pm$3.90 & 74.83$\pm$3.12 & 75.45$\pm$3.26 & 71.25$\pm$4.43 & 72.88$\pm$2.05 & 75.49$\pm$2.03 & 75.86$\pm$2.30 & 73.70$\pm$5.70  & \textbf{78.24$\pm$4.37}\\       			Seeds & 87.22$\pm$7.08 & 91.93$\pm$2.58 & 92.83$\pm$1.76 & 90.63$\pm$2.40 & 90.42$\pm$2.83 & 78.31$\pm$20.88 & 89.89$\pm$0.69 & 91.48$\pm$1.60 & 89.47$\pm$2.93 & 92.06$\pm$0.63  & \textbf{93.17$\pm$1.12}\\       			glass & 64.26$\pm$7.54 & 77.16$\pm$4.27 & 73.58$\pm$4.79 & 78.74$\pm$4.18 & 77.85$\pm$4.09 & 56.39$\pm$20.35 & 71.47$\pm$5.92 & \textbf{79.63$\pm$3.20} & 76.34$\pm$5.79 & 74.55$\pm$5.49  & 74.87$\pm$5.47\\       			zoo   & 78.78$\pm$10.07 & \textbf{98.58$\pm$1.24} & 97.03$\pm$1.99 & 92.59$\pm$3.57 & 92.82$\pm$2.65 & 85.06$\pm$4.34 & 91.06$\pm$3.12 & 92.59$\pm$3.41 & 92.24$\pm$2.94 & 93.06$\pm$3.30  & 93.06$\pm$1.11\\       			appendicitis & 81.44$\pm$5.92 & 75.11$\pm$11.76 & 72.55$\pm$5.83 & 80.43$\pm$7.86 & 80.32$\pm$7.81 & 81.17$\pm$5.81 & 82.34$\pm$2.29 & 82.66$\pm$7.93 & 81.49$\pm$6.83 & 83.19$\pm$6.86  & \textbf{83.30$\pm$5.69}\\       			heart & 71.71$\pm$6.32 & 73.27$\pm$4.16 & 73.85$\pm$3.18 & 76.09$\pm$5.46 & 76.13$\pm$4.12 & 75.51$\pm$9.46 & 79.14$\pm$3.39 & 76.67$\pm$6.58 & 77.28$\pm$6.13 & 81.11$\pm$1.31  & \textbf{81.65$\pm$1.38}\\       			cleve & 71.07$\pm$5.87 & 73.98$\pm$5.49 & 71.60$\pm$3.15 & 76.28$\pm$2.81 & 75.79$\pm$3.63 & 71.50$\pm$4.59 & 76.39$\pm$1.48 & 76.88$\pm$2.98 & 77.03$\pm$3.79 & 77.37$\pm$1.93  & \textbf{77.44$\pm$2.35}\\       			segmentation & 55.63$\pm$5.85 & 76.06$\pm$6.44 & 80.61$\pm$5.02 & 76.98$\pm$2.91 & 75.98$\pm$3.55 & 69.21$\pm$7.27 & 71.32$\pm$3.03 & 78.20$\pm$3.84 & 74.66$\pm$3.31 & 76.51$\pm$4.13  & \textbf{79.10$\pm$4.16}\\       			X8D5K & 94.33$\pm$4.29 & 100.0$\pm$0.0 & 100.0$\pm$0.0 & 100.0$\pm$0.0 & 100.0$\pm$0.0 & 100.0$\pm$0.0 & 100.0$\pm$0.0 & 100.0$\pm$0.0 & 100.0$\pm$0.0 & \textbf{100.0$\pm$0.0}  & \textbf{100.00$\pm$0.00}\\       			vote  & 80.54$\pm$5.32 & 90.52$\pm$1.75 & 84.90$\pm$3.02 & 89.23$\pm$2.48 & 89.64$\pm$2.42 & 87.37$\pm$2.50 & 87.34$\pm$1.04 & 90.20$\pm$2.06 & 89.85$\pm$2.19 & 91.10$\pm$1.51  & \textbf{91.79$\pm$1.46}\\       			WBC   & 92.73$\pm$3.43 & 92.38$\pm$2.68 & 93.98$\pm$1.33 & 95.70$\pm$1.56 & 95.88$\pm$1.07 & \textbf{96.69$\pm$0.51} & 95.90$\pm$0.51 & 95.64$\pm$0.50 & 95.65$\pm$0.81 & 96.19$\pm$0.47  & 96.48$\pm$0.66\\       			breast & 69.96$\pm$3.86 & 63.47$\pm$4.85 & 71.73$\pm$2.52 & 66.25$\pm$4.18 & 66.90$\pm$4.05 & 71.13$\pm$3.29 & 70.97$\pm$0.00 & 67.50$\pm$3.13 & 68.79$\pm$3.07 & 72.10$\pm$1.41  & \textbf{73.35$\pm$1.49}\\       			austra & 78.54$\pm$5.48 & 77.65$\pm$3.83 & 81.49$\pm$2.37 & 85.44$\pm$1.64 & 85.23$\pm$1.98 & 76.97$\pm$11.37 & 81.87$\pm$1.37 & 84.35$\pm$1.06 & 83.98$\pm$3.10 & 85.16$\pm$1.54  & \textbf{85.66$\pm$0.68}\\       			diabetes & 69.93$\pm$2.96 & 66.75$\pm$2.75 & \textbf{70.69$\pm$1.55} & 68.13$\pm$0.92 & 69.10$\pm$1.05 & 67.13$\pm$1.81 & 65.01$\pm$0.87 & 70.09$\pm$1.13 & 69.10$\pm$1.50 & 69.31$\pm$1.55  & 69.48$\pm$1.10\\      			dnatest & 48.96$\pm$2.27 & 82.95$\pm$1.53 & 83.47$\pm$1.19 & 85.30$\pm$1.07 & 85.54$\pm$1.05 & 81.91$\pm$1.42 & 74.40$\pm$1.85 & \textbf{87.80$\pm$1.00} & 82.03$\pm$5.27 & 84.92$\pm$0.92  & 85.39$\pm$1.27\\       			\midrule      			Average & 75.29$\pm$5.20 & 81.37$\pm$3.84 & 80.84$\pm$3.01 & 82.86$\pm$3.16 & 83.21$\pm$3.15 & 80.04$\pm$5.39 & 81.39$\pm$1.92 & 83.88$\pm$2.64 & 83.05$\pm$3.37 & 84.62$\pm$1.91 & \textbf{85.27$\pm$1.71}\\  			
\bottomrule      		
\end{tabular} } 
\end{table*} \begin{table*}[!htp] 	 
\centering 
\caption{Summary of ranks computed by the Wilcoxon test. "$\shortuparrow$" denotes the method in the row improves the method of the column. "$\shortdownarrow$" indicates the method in the column improves the method of the row. The upper diagonal of level significance is 0.1 while the lower diagonal level of significance is 0.05.} 
\label{tab:ranks}
\begin{tabular}{rrrrrrrrrrrr} 
\toprule 		 
& SS-ELM   & TSVM     & LapRLS  & GCN-Linear & GCN      & DropEdge & SGC      & GCNII   & APPNP    & FLGC   & FLGC* \\  
\midrule 		 
SS-ELM     & -        & 66.0 $\shortdownarrow$  & 54.0 $\shortdownarrow$ & 23.0 $\shortdownarrow$    & 18.0 $\shortdownarrow$  & 56.0 $\shortdownarrow$  & 15.0 $\shortdownarrow$  & 8.0 $\shortdownarrow$  & 10.0 $\shortdownarrow$  & 1.0 $\shortdownarrow$  & 1.0 $\shortdownarrow$ \\ 		 
TSVM       & 234.0 $\shortuparrow$ & -        & 136.0   & 74.0 $\shortdownarrow$    & 66.0 $\shortdownarrow$  & 157.0    & 142.0    & 48.0 $\shortdownarrow$ & 80.0 $\shortdownarrow$  & 53.0 $\shortdownarrow$ & 37.0 $\shortdownarrow$ \\ 	 
LapRLS     & 246.0 $\shortuparrow$ & 140.0    & -       & 110.0      & 100.0    & 167.0    & 146.0    & 74.0 $\shortdownarrow$ & 106.0    & 67.0 $\shortdownarrow$ & 41.0 $\shortdownarrow$ \\ 		
GCN-Linear & 277.0 $\shortuparrow$ & 202.0    & 166.0   & -          & 99.0     & 215.0 $\shortuparrow$ & 191.0    & 42.5 $\shortdownarrow$ & 135.5    & 41.5 $\shortdownarrow$ & 22.0 $\shortdownarrow$ \\ 		 GCN        & 282.0 $\shortuparrow$ & 210.0 $\shortuparrow$ & 176.0   & 177.0      & -        & 231.0 $\shortuparrow$ & 197.0 $\shortuparrow$ & 48.0 $\shortdownarrow$ & 172.5    & 63.0 & 38.0 $\shortdownarrow$ \\ 		 DropEdge   & 244.0 $\shortuparrow$ & 119.0    & 109.0   & 61.0 $\shortdownarrow$    & 45.0 $\shortdownarrow$  & -        & 101.0    & 34.0 $\shortdownarrow$ & 37.0 $\shortdownarrow$  & 13.0 $\shortdownarrow$ & 9.0 $\shortdownarrow$ \\ 		 SGC        & 285.0 $\shortuparrow$ & 134.0    & 130.0   & 85.0       & 79.0     & 175.0    & -        & 65.5 $\shortdownarrow$ & 71.0 $\shortdownarrow$  & 5.0 $\shortdownarrow$  & 5.0 $\shortdownarrow$ \\ 		 GCNII      & 292.0 $\shortuparrow$ & 228.0 $\shortuparrow$ & 202.0   & 257.5 $\shortuparrow$   & 228.0 $\shortuparrow$ & 242.0 $\shortuparrow$ & 234.5 $\shortuparrow$ & -       & 208.0 $\shortuparrow$ & 104.5   &  65.5 $\shortdownarrow$ \\ 		 APPNP      & 290.0 $\shortuparrow$ & 196.0    & 170.0   & 140.5      & 127.5    & 239.0 $\shortuparrow$ & 205.0 $\shortuparrow$ & 68.0 $\shortdownarrow$ & -        & 46.0 $\shortdownarrow$ & 29.0 $\shortdownarrow$ \\ 		 FLGC       & 299.0 $\shortuparrow$ & 223.0 $\shortuparrow$ & 209.0$\shortuparrow$ & 235.5 $\shortuparrow$   & 214.0 $\shortuparrow$ & 263.0 $\shortuparrow$ & 295.0 $\shortuparrow$ & 195.5   & 230.0 $\shortuparrow$ & -   & 5.0 $\shortdownarrow$ \\ FLGC* & 299.0 $\shortuparrow$ & 239.0 $\shortuparrow$ & 235.0 $\shortuparrow$ & 254.0 $\shortuparrow$ & 238.0 $\shortuparrow$ & 267.0 $\shortuparrow$ & 295.0 $\shortuparrow$ & 234.5 $\shortuparrow$ & 247.0 $\shortuparrow$ & 271.0 $\shortuparrow$ & -\\  
\bottomrule 	 
\end{tabular} 
\end{table*}
\begin{figure*}[tbh]
\subfloat[]{\includegraphics[width=0.66\columnwidth]{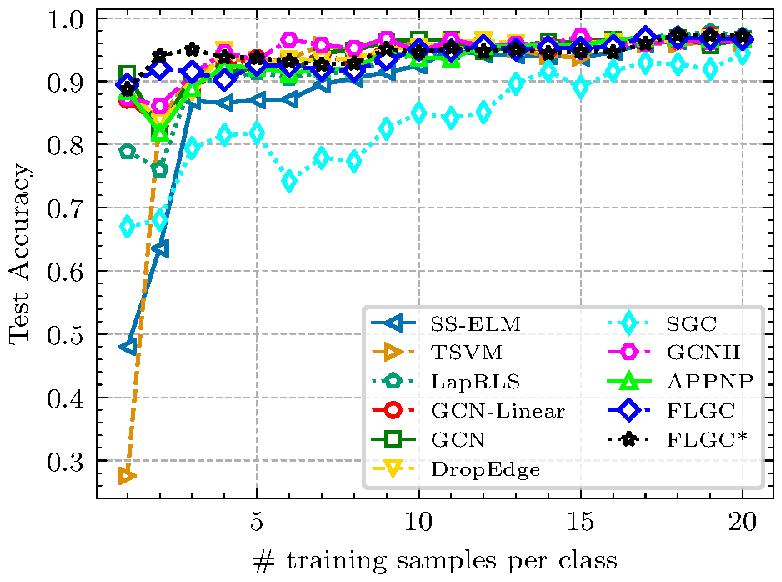}}\subfloat[]{\includegraphics[width=0.66\columnwidth]{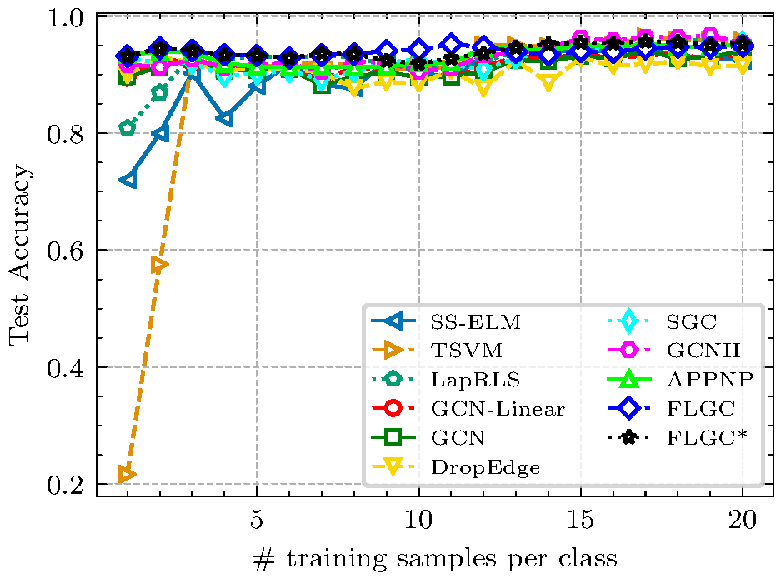}}\subfloat[]{\includegraphics[width=0.66\columnwidth]{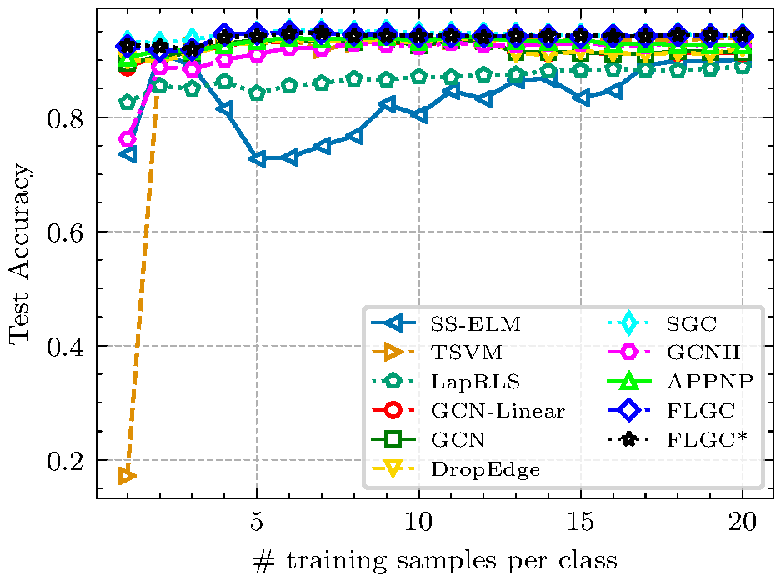}}

\caption{Test accuracy with different training samples per class on (a) Iris,
(b) Wine, (c) WDBC datasets.\label{fig:acc-samples}}
\end{figure*}
\begin{figure*}[tbh]
\subfloat[]{\includegraphics[width=0.66\columnwidth]{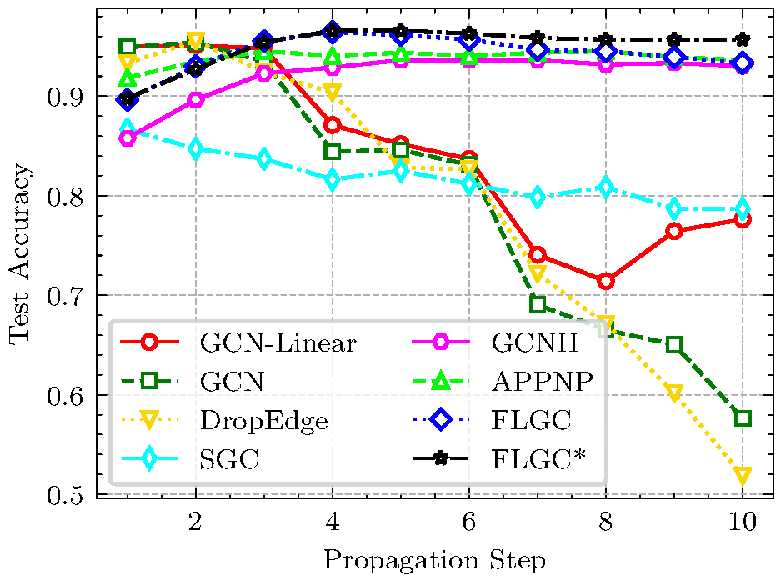}}\subfloat[]{\includegraphics[width=0.66\columnwidth]{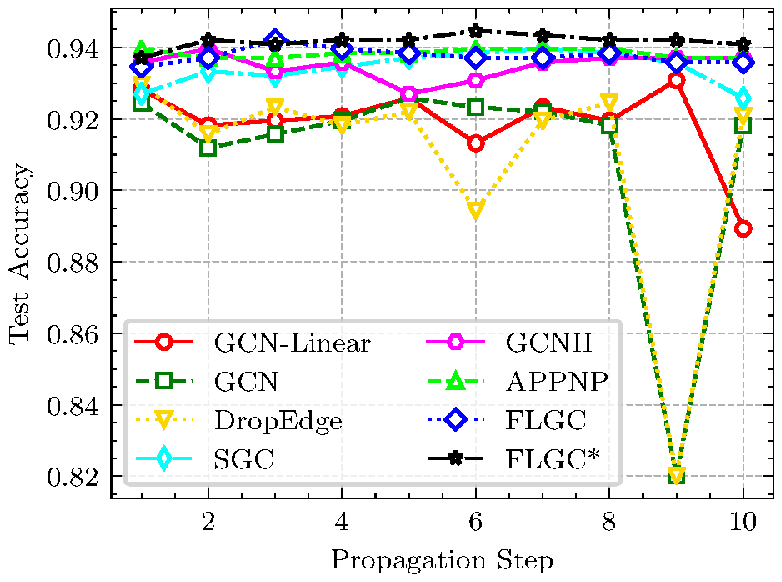}}\subfloat[]{\includegraphics[width=0.66\columnwidth]{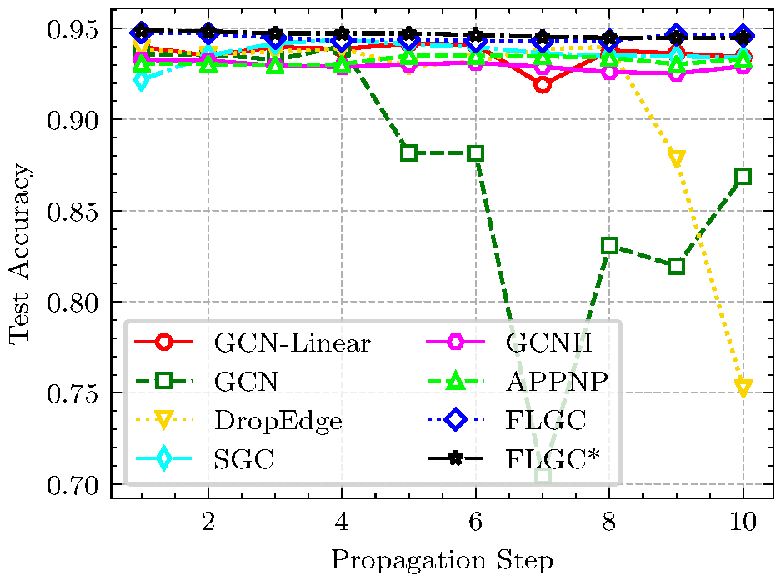}}

\caption{Over-smoothing effect on (a) Iris, (b) Wine, (c) WDBC datasets. \label{fig:over-smooth-uci}}
\end{figure*}
\begin{figure*}[tbh]
\subfloat[]{\includegraphics[width=0.66\columnwidth]{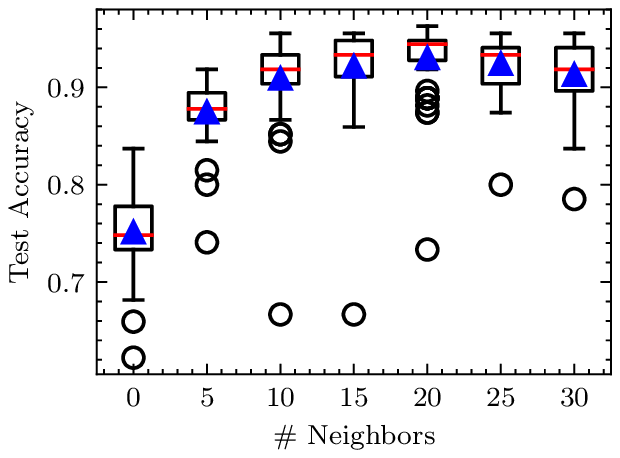}}\subfloat[]{\includegraphics[width=0.66\columnwidth]{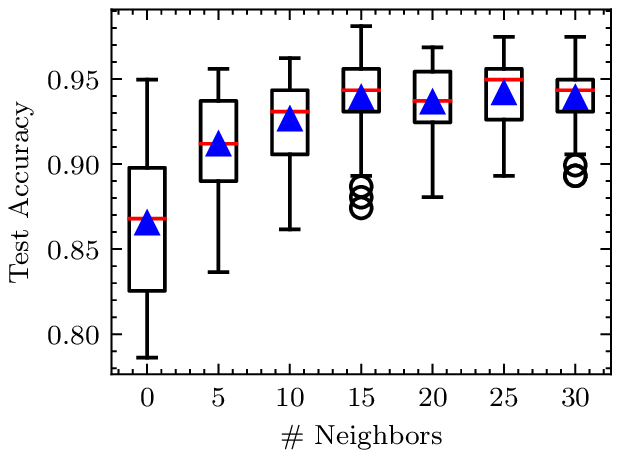}}\subfloat[]{\includegraphics[width=0.66\columnwidth]{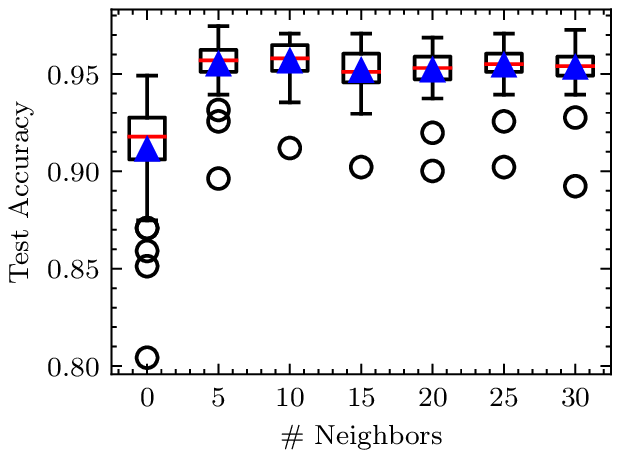}}

\caption{Effect of neighbor size on (a) Iris, (b) Wine, (c) WDBC datasets.
We perform FLGC by fixing $\lambda=10^{-4}$, $K=5$, and $\alpha=0.1$
and varying $k\in\left\{ 0,5,10,15,20,25,30\right\} $. Noticed that
$k=0$ means $\mathbf{A}=\mathbf{I}$ and thus only $\mathbf{X}$
is used. Each box indicates the statistic over $50$ independent runs
and corresponding average value is denoted as a blue triangle. \label{fig:neigh-box-uci}}
\end{figure*}

\subsubsection{Baselines and Setup}

In this experiment, the selected baselines include GCN variants, i.e.,
GCN-Linear, GCN \cite{GCN-kipf-ICLR-2017}, DropEdge \cite{DropEdge-ICLR-2019},
SGC \cite{GCN-simplifying-wu-ICML-2019}, GCNII \cite{GCNII-Chen-ICML-2020},
and APPNP \cite{APPNP-klicpera-ICLR-2019}, and classic semi-supervised
classification models, i.e., SS-ELM \cite{Semi-Supervised-and-Unsupervised-ELM-Huang-TCYB-2014},
TSVM \cite{Semi-TSVM-Joachims-ICML-1999}, and LapRLS \cite{Semi-LapRLS-LapSVM-JMLR-2006}.
There is no off-the-shelf structure in these regular datasets, thus,
we construct $k$NN graphs \cite{HSI-Clustering-GCSC-CAI-TGRS-2020,HyperAE-CAI-GRSL-2021}
for the representation of structured information. Specifically, we
adopt the Euclidean distance to measure the similarity between sample
pairs and choose top $k$ neighbors centered on a certain sample as
its edges. To avoid hyper-parameter $k$, we empirically set it as
$k=\left\lfloor \frac{N}{5C}\right\rfloor $. 

\subsubsection{Comparison with SOTAs and Statistical Test}

In Table \ref{tab:acc-res-uci}, we provide the comparative results
on the 24 UCI datasets. All the results are calculated by averaging
$30$ independent runs. At the bottom of the table, we summarize the
arithmetic mean accuracy over 24 datasets. Remarkably, our FLGC models
consistently outperform not only the classic semi-supervised models
but also the recent GCN variants. Specifically, FLGC{*} achieves the
highest accuracy on $16$ out of $24$ datasets. On average, FLGC
and FLGC{*} respectively obtain $84.62\pm1.91$ and $85.27\pm1.71$
accuracy across 24 datasets, which improve upon SGC and APPNP by margins
of $3.23\%$ and $2.22\%$, respectively. Furthermore, we notice that
GCN variants are generally superior to classic semi-supervised models
even on the regular grid datasets. This is a valuable clue in designing
semi-supervised models on regular datasets.

To further rank all baselines, we carry on a non-parametric statistical
test on the results reported in Table \ref{tab:acc-res-uci}. To this
end, we follow the suggestion posed by Garcia \emph{et al.} \cite{Wilcoxon-stat-compar-Garcia-JMLR-2008}
on adopting the Wilcoxon signed-ranks test\footnote{We use KEEL (Knowledge Extraction based on Evolutionary Learning)
tool available from http://www.keel.es/ to conduct the Wilcoxon signed-ranks
test.} to compute the sum of ranks for each pair of methods. Table \ref{tab:ranks}
shows the detailed statistical results. According to the exact critical
value table of the Wilcoxon test, the critical values on $24$ datasets
for a confidence level of $0.05$ and $0.1$ correspond to $81$ (lower
diagonal) and $91$ ( upper diagonal), respectively. We can observe
that our proposed FLGC{*} is significantly better than all the other
competitors for different confidence levels, while the FLGC model
performs equally to GCNII and GCN for the confidence level of $0.05$
and $0.1$, respectively. The results markedly demonstrate that our
proposed FLGC models can generalize to the regular Euclidean data
and can achieve promising performance.

\subsubsection{Comparison w.r.t. Different Sizes of Training Samples}

We conduct an experiment to compare the performance of baselines with
varying labeled samples size. Fig. \ref{fig:acc-samples} (a)-(c)
illustrate the comparative results on three selected datasets, i.e.,
Iris, Wine, and WDBC. We gradually increase the training samples per
class from 1 to 20 and plot the test accuracy curves of $11$ competitive
methods. It can be seen that both of our FLGC models show competitive
performance w.r.t. the other baselines under different training sizes.
Particularly, our methods remarkably outperform many baselines (e.g.,
SS-ELM, TSVM, and LapRLS) using all datasets when using an extremely
small training size, e.g., only 1 labeled sample per class. 

\subsubsection{Study on Over-Smoothing}

Fig. \ref{fig:over-smooth-uci} (a)-(c) show the effect of the propagation
step on the selected datasets. Several tendencies can be observed
from the figures. Firstly, GCNII, APPNP, FLGC, and FLGC{*} face less
risk of falling into the over-smoothing dilemma, while that occurs
in the other methods. Secondly, linear GCN tends to outperform the
nonlinear GCN on the selected datasets. A conceivable reason is that
the nonlinear activation accelerates the speed of over-smoothing.
Also, this is affected by the predefined graph structure. Thirdly,
by adding initial residual, FLGC{*} benefits from the longer-range
propagation, and thus significantly improves upon FLGC.
\begin{figure*}[tbh]
\begin{centering}
\subfloat[]{\includegraphics[width=0.66\columnwidth]{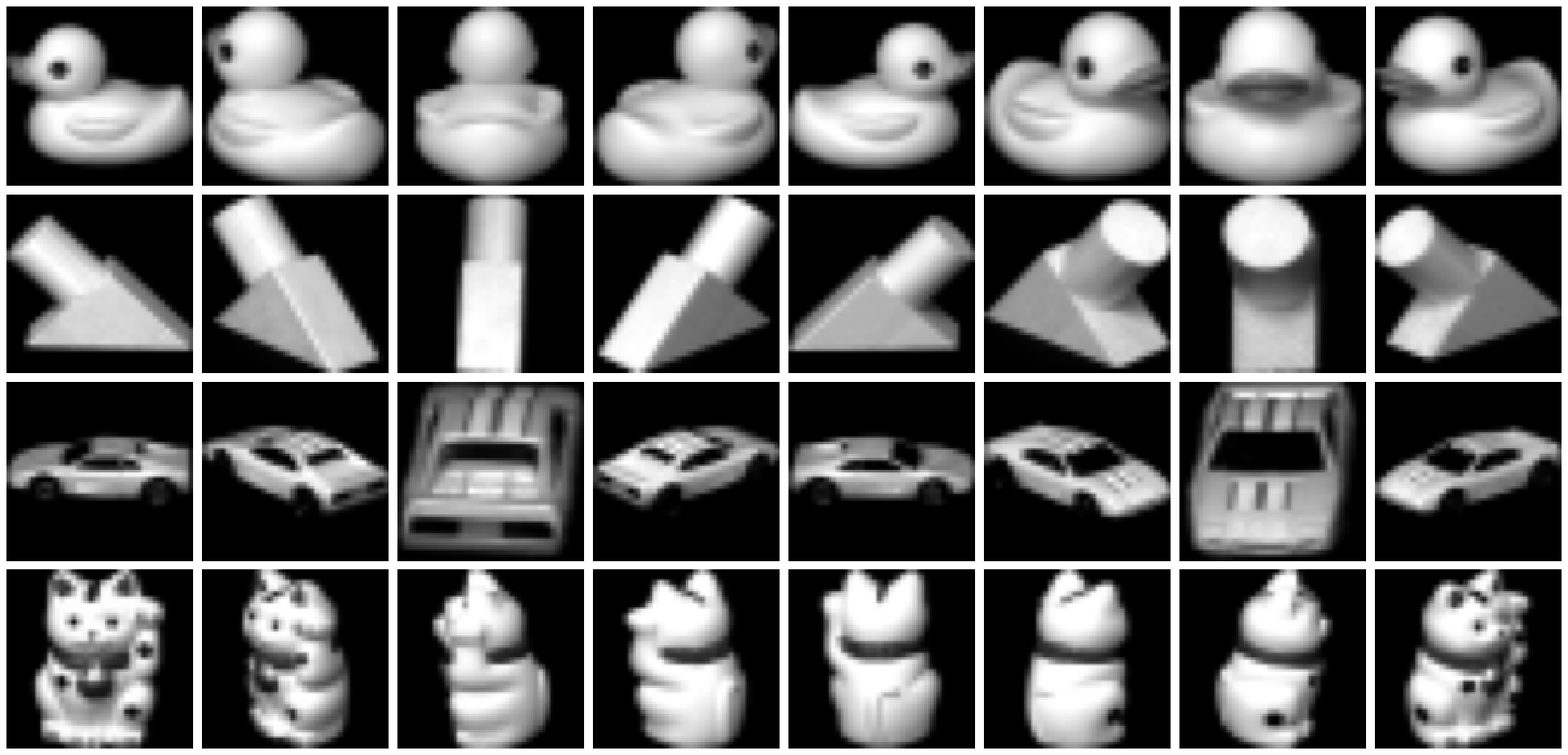}}\subfloat[]{\includegraphics[width=0.66\columnwidth]{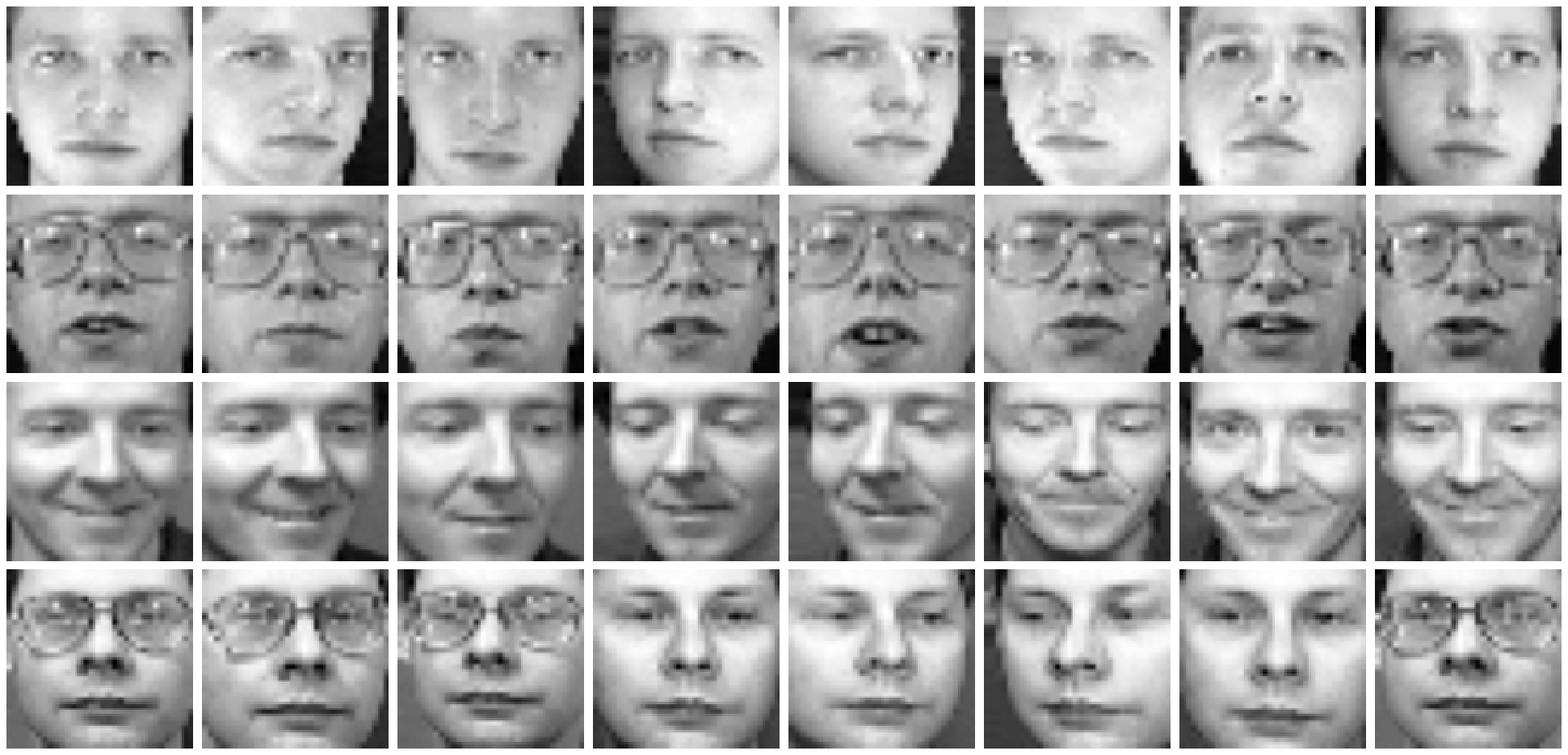}}\subfloat[]{\includegraphics[width=0.66\columnwidth]{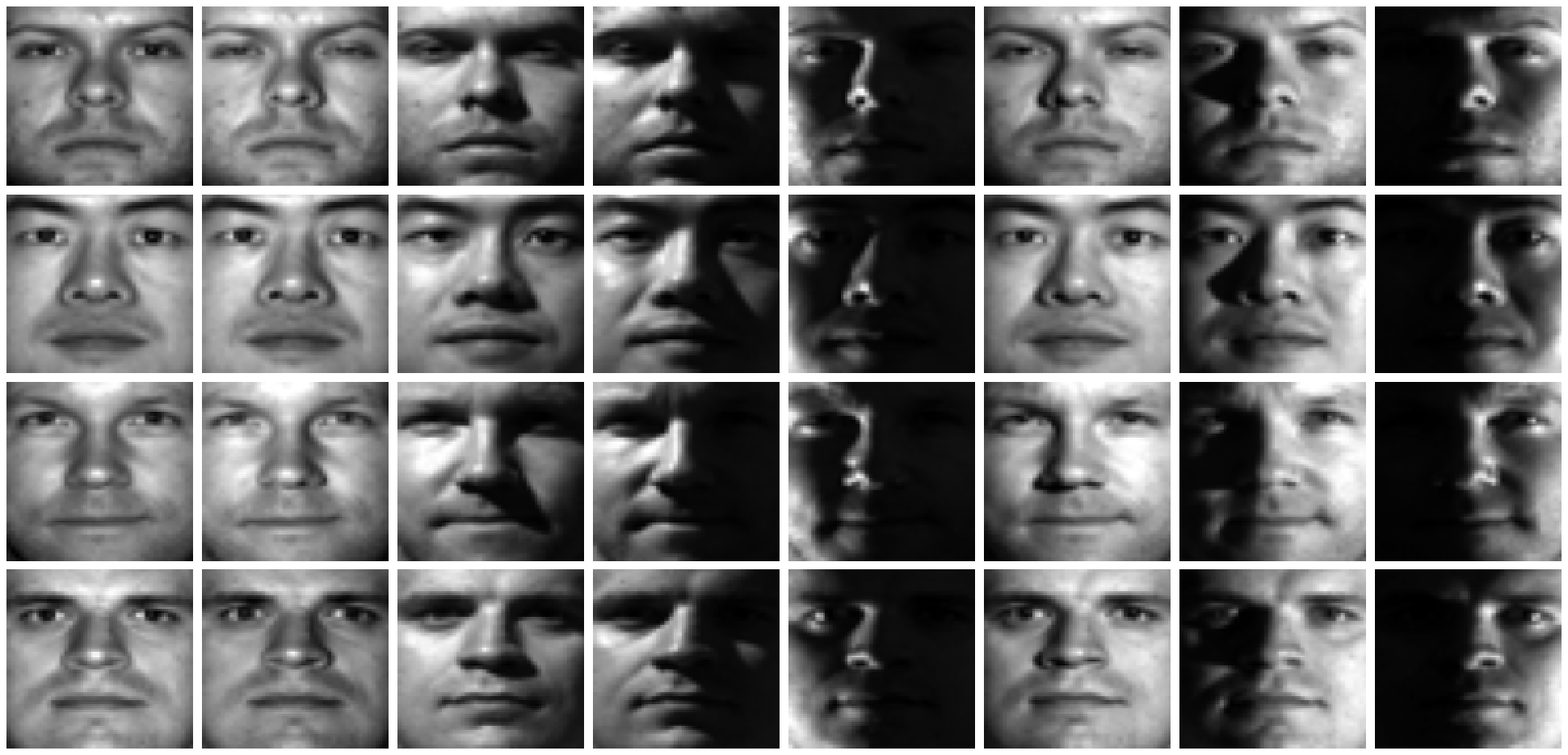}}
\par\end{centering}
\caption{Sample images from (a) COIL20, (b) ORL, and (c) Extended Yale B.\label{fig:Sample-images-face}}
\end{figure*}
\begin{table*}[htbp]   
\centering   
\caption{The clustering performance of different clustering models on Iris, Wine, and Yale datasets.}     
\begin{tabular}{cc|ccccccccccc}     
\toprule     
Datasets & Metric & K-Means & NMF   & NCut  & CAN   & CLR   & SSR   & KMM   & RRCSL & EGCSC & FLGC  & FLGC* \\     
\midrule     
\multirow{2}[2]{*}{Iris} & ACC   & 0.797 & 0.636 & 0.586 & 0.573 & 0.846 & 0.686 & 0.817 & 0.873 & 0.940  & 0.953 & \textbf{0.960} \\           & NMI   & 0.637 & 0.404 & 0.572 & 0.427 & 0.723 & 0.580  & 0.682 & 0.754 & 0.851 & 0.874 & \textbf{0.886} \\     
\midrule     
\multirow{2}[2]{*}{Wine} & ACC   & 0.960  & 0.722 & 0.959 & 0.926 & 0.915 & 0.955 & 0.541 & 0.972 & 0.949 & 0.949 & \textbf{0.983} \\           & NMI   & 0.862 & 0.447 & 0.881 & 0.780  & 0.732 & 0.848 & 0.361 & 0.881 & 0.820  & 0.820  & \textbf{0.928} \\     
\midrule     
\multirow{2}[2]{*}{Yale} & ACC   & 0.472 & 0.339 & 0.511 & 0.521 & 0.509 & 0.593 & 0.442 & 0.600   & 0.515 & 0.546 & \textbf{0.630} \\           & NMI   & 0.540  & 0.412 & 0.561 & 0.549 & 0.582 & 0.584 & 0.506 & 0.631 & 0.558 & 0.557 & \textbf{0.657} \\     
\bottomrule     
\end{tabular}   
\label{tab:acc-nmi-small}
\end{table*}
\begin{figure*}[tbh]
\begin{centering}
\subfloat[]{\includegraphics[width=0.66\columnwidth]{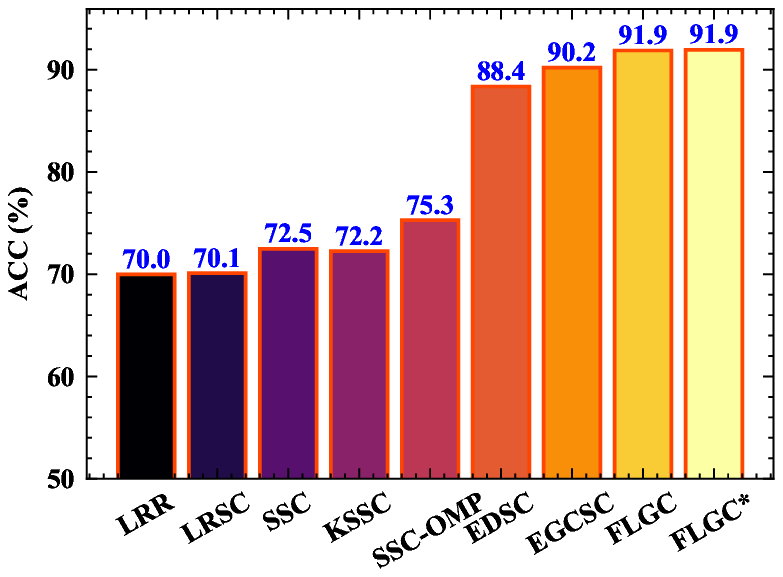}

}\subfloat[]{\includegraphics[width=0.66\columnwidth]{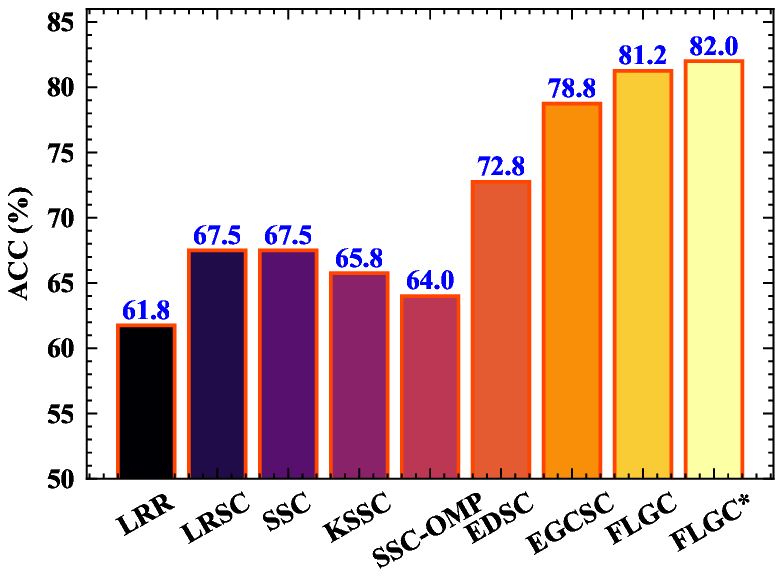}

}\subfloat[]{\includegraphics[width=0.66\columnwidth]{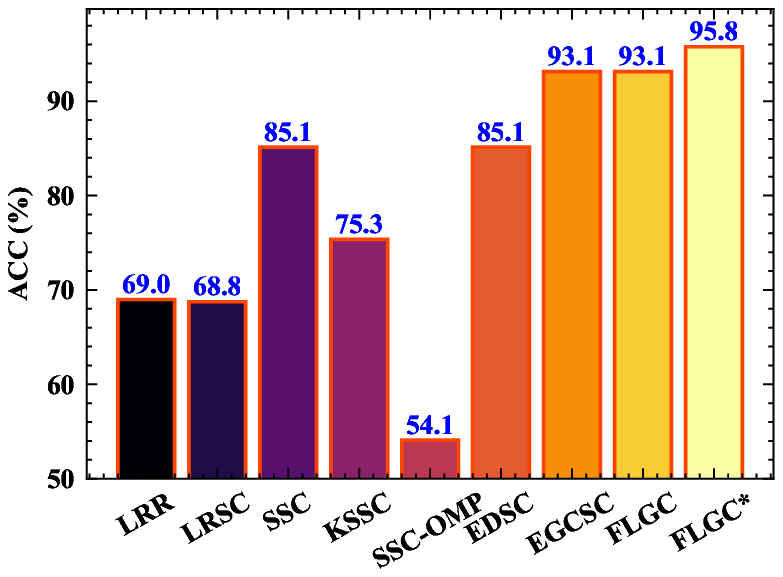}

}
\par\end{centering}
\caption{Clustering ACC (in \%) on the (a) COIL20, (b) ORL, and (c) Extended
Yale B datasets. Different colors indicate different methods. The
height of the bars indicates the clustering ACC, so the higher the
better. \label{fig:Clustering-ACC-Bar}}
\end{figure*}
\begin{figure}[tbh]
\begin{centering}
\subfloat[]{\includegraphics[width=0.5\columnwidth]{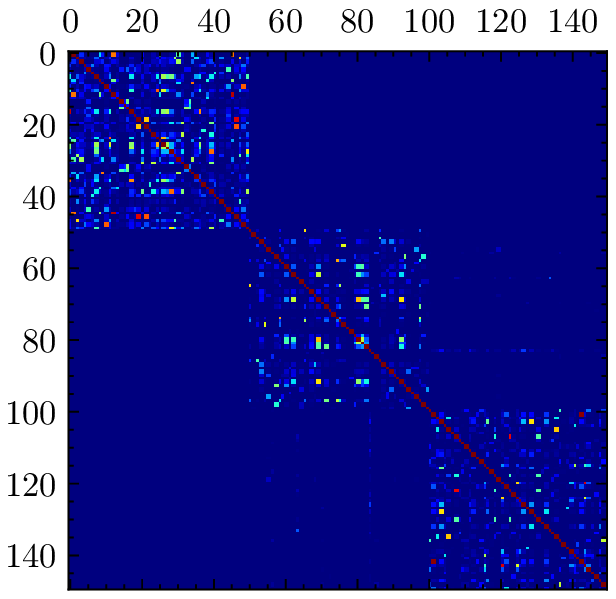}}\subfloat[]{\includegraphics[width=0.5\columnwidth]{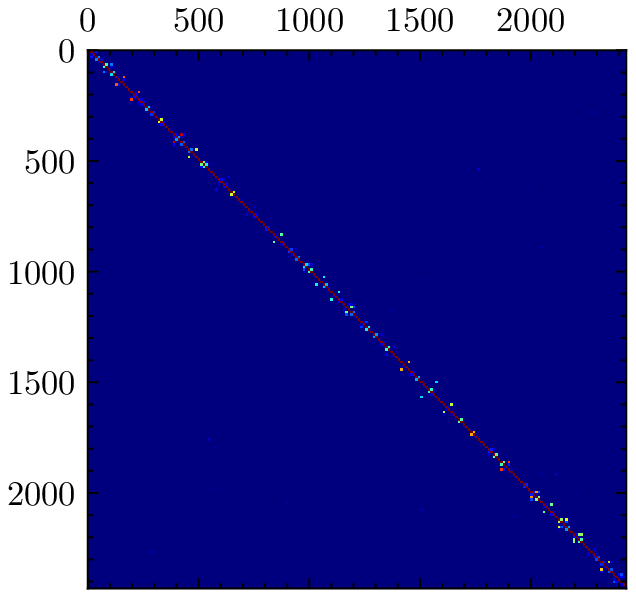}}
\par\end{centering}
\caption{Affinity matrix obtained by FLGC{*} for (a) Iris and (b) Extended
Yale B datasets.\label{fig:Affinity}}
\end{figure}
\begin{table*}[htbp]   
\centering   
\caption{Clustering results of the robustness experiments on the Extended Yale B dataset.} 
\label{tab:robust}   
\begin{tabular}{c|lcccccccc}     
\toprule     
\multicolumn{1}{c}{\multirow{2}[4]{*}{Noise Type}} & \multirow{2}[4]{*}{Noise Intensity} & \multicolumn{4}{c}{ACC}       & \multicolumn{4}{c}{NMI} \\ \cmidrule(r){3-6} \cmidrule(r){7-10}   \multicolumn{1}{c}{} &       & \multicolumn{1}{c}{EDSC} & \multicolumn{1}{c}{EGCSC} & \multicolumn{1}{c}{FLGC} & \multicolumn{1}{c}{FLGC*} & \multicolumn{1}{c}{EDSC} & \multicolumn{1}{c}{EGCSC} & \multicolumn{1}{c}{FLGC} & \multicolumn{1}{c}{FLGC*} \\     \midrule     \multirow{4}[2]{*}{Gaussian} & $\sigma^2=0.01$ & 0.853 & 0.863 & 0.863 & \textbf{0.901} & 0.897 & 0.881 & 0.881 & \textbf{0.919} \\           & $\sigma^2=0.05$ & 0.838 & 0.830  & 0.830  & \textbf{0.891} & 0.884 & 0.855 & 0.855 & \textbf{0.906} \\           & $\sigma^2=0.1$ & 0.472 & 0.748 & 0.748 & \textbf{0.824} & 0.748 & 0.780  & 0.780  & \textbf{0.866} \\           & $\sigma^2=0.2$ & 0.680  & 0.643 & 0.643 & \textbf{0.774} & 0.774 & 0.710  & 0.710  & \textbf{0.813} \\ \cmidrule{1-10}    
\multirow{4}[2]{*}{Salt \& Pepper} & $p=0.01$ & 0.845 & 0.891 & 0.891 & \textbf{0.908} & 0.886 & 0.901 & 0.901 & \textbf{0.922} \\           & $p=0.05$ & 0.842 & 0.862 & 0.862 & \textbf{0.896} & 0.885 & 0.883 & 0.883 & \textbf{0.913} \\           & $p=0.1$ & 0.843 & 0.847 & 0.847 & \textbf{0.894} & 0.885 & 0.870  & 0.870  & \textbf{0.912} \\           & $p=0.2$ & 0.838 & 0.826 & 0.826 & \textbf{0.877} & 0.881 & 0.850  & 0.850  & \textbf{0.901} \\     
\bottomrule     
\end{tabular} 
\label{tab:addlabel}
\end{table*}
\begin{figure*}[tbh]
\begin{centering}
\subfloat[]{\includegraphics[width=0.66\columnwidth]{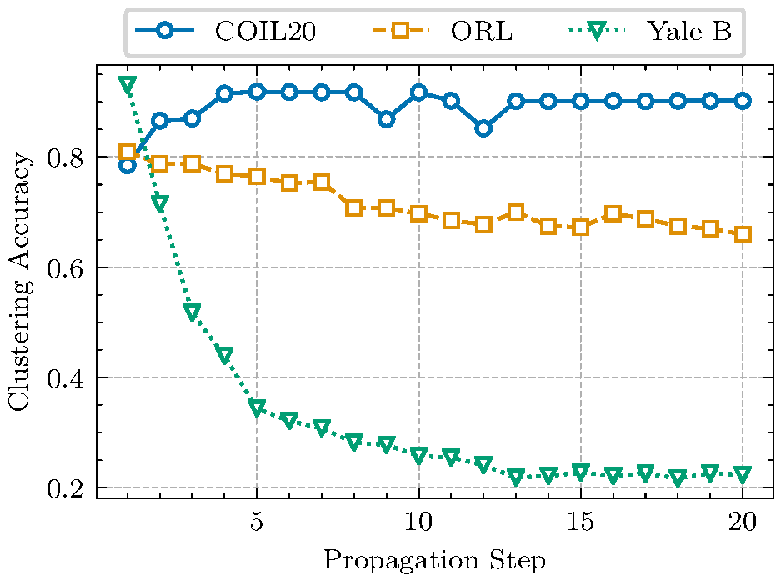}

}\subfloat[]{\includegraphics[width=0.66\columnwidth]{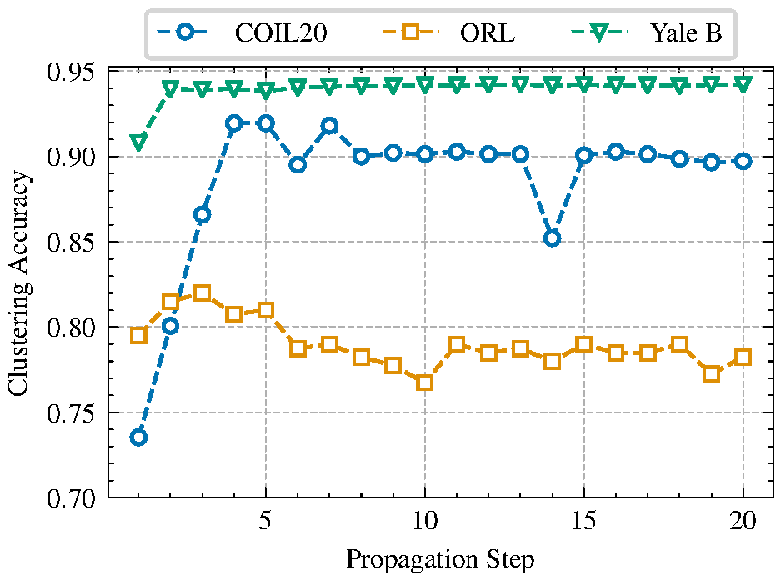}

}\subfloat[]{\includegraphics[width=0.66\columnwidth]{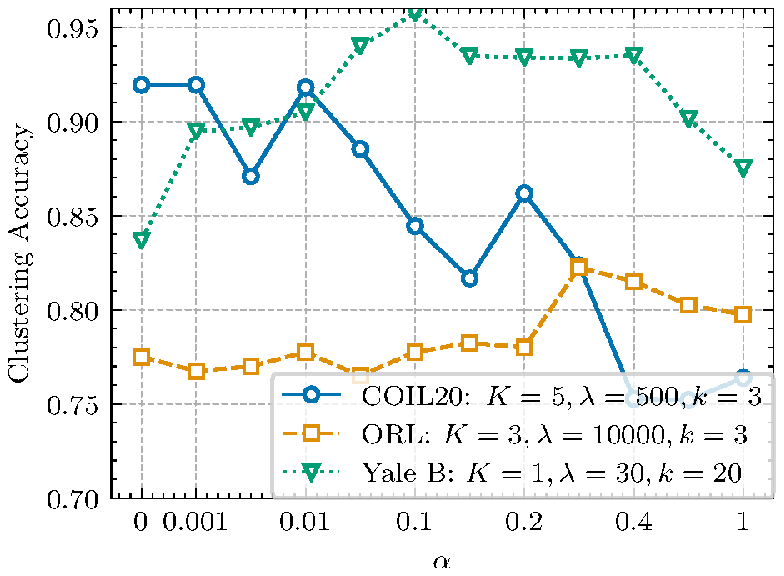}

}
\par\end{centering}
\caption{Effect of propagation step on the clustering accuracy of (a) FLGC
and (b) FLGC{*}. (c) The influence of $\alpha$ on three benchmarks.
\label{fig:propagation-clu}}
\end{figure*}

\subsubsection{Effect of the Neighborhood Size}

We aim to further explore the effect of the predefined graph structure
on the classification performance of our FLGC. Fig. \ref{fig:neigh-box-uci}
shows the tendency of accuracy varied with neighborhood sizes. When
$k=0$, meaning $\mathbf{P}=\mathbf{I}$, FLGC performs identically
to a ridge regression classifier. At this point, FLGC's performance
becomes more unstable and worse than that considered neighborhood
information. This demonstrates that a pairwise relationship defines
the intrinsic structure of regular data. Different from real-word
graph data, however, the predefined edges (i.e., $k$NN graph) cannot
perfectly describe such structures. As a result, the performance of
FLGC varies with neighborhood sizes. Empirically, a large neighborhood
leads to more performance improvement since that enlarges the first-order
receptive field. However, a too large neighborhood size will inevitably
degrade the performance because of the increased risk of noisy edges
and over-smoothing. It is still an open problem to find an optimum
neighborhood size.

\section{Experiments of Clustering \label{sec:Experiments-of-Clustering}}

In this section, we extensively evaluate our proposed unsupervised
FLGC on several challenging clustering benchmarks and compare it with
many previous clustering models.

\subsection{Dataset Description}

Except for two simple UCI datasets (i.e., Iris and Wine), we add four
challenging image clustering benchmarks for performance comparison,
i.e., COIL20 object image dataset and Yale, Extended Yale B, and ORL
face image datasets. The COIL20 dataset contains $1,440$ gray-scale
image samples with a spatial size of $32\times32$ and consists of
$20$ distinct objects, e.g., duck, cat, and car models. The ORL dataset
is composed of $400$ human face images, with $40$ subjects each
having $10$ samples. Following \cite{DeepSC-JPan-NIPS2017}, we down-sampled
the original face images from $112\times92$ to $32\times32$. The
Yale and Extended Yale B datasets are popular benchmarks for subspace
clustering. The former includes $165$ face images collected from
$15$ individuals. The latter database is more challenging than the
former because it contains $2,432$ images of $38$ human subjects
acquired under different poses and illumination conditions. The resolution
of the two face databases is scaled to $32\times32$ and $42\times48$,
respectively. Some selected sample images from COIL20, ORL, and Extended
Yale B are illustrated in Fig. \ref{fig:Sample-images-face}.

\subsection{Baselines and Setup}

We divide our experiment into two parts. In the first part, we aim
to evaluate our methods on three small and simple datasets, i.e.,
Iris, Wine, and Yale. This part follows the settings suggested in
\cite{RRCSL-TCYB-WANG-2021}. We compare our method with the following
baselines: K-Means, NMF \cite{NMF-Lee-Nature-1999}, Normalized Cuts
(NCut) \cite{NCut-SHI-TPAMI-2000}, CAN \cite{CAN-Nie-KDD-2014},
CLR \cite{CLR-NIe-AAAI-2016}, SSR \cite{SSR-Huang-IJCAI-15}, K-Multiple-Means
(KMM) \cite{KMMs-Nie-KDD-19}, EGCSC \cite{HSI-Clustering-GCSC-CAI-TGRS-2020},
and RRCSL \cite{RRCSL-TCYB-WANG-2021}. Similar to Section \ref{sec:Experiments-of-Traductive},
we use FLGC{*} to denote our method with the initial residual. In
the second part, we conduct experiments on three challenging image
datasets (i.e., COIL20, ORL, and Extended Yale B). The baselines that
we compare FLGCs with in this part include Low Rank Representation
(LRR) \cite{LRR-TPAMI-Liu-2012}, LRSC \cite{LRSC-1-Vidal-PRLetters-2014},
SSC \cite{SSC-Elhamifar-TPAMI-2013}, KSSC \cite{KSSC-Patel-ICIP-2017},
SSC by Orthogonal Matching Pursuit (SSC-OMP) \cite{OMP-You-CVPR-2016},
EDSC \cite{EDSC-PanJ-WACV-2014}, and EGCSC \cite{HSI-Clustering-GCSC-CAI-TGRS-2020}.
We follow the experiment setups reported in \cite{DeepSC-JPan-NIPS2017}.
For our FLGC, we search for the optimum parameter setting among $\lambda\in\left[1,10^{4}\right]$,
$K\in\left[1,20\right]$, $\alpha\in\left[0,1\right]$, and a $k$NN
graph of $k=\left\lfloor \frac{N}{5C}\right\rfloor $.

Two popular metrics are utilized to quantify the performance of the
clustering methods, including clustering Accuracy (ACC) and Normalized
Mutual Information (NMI) \cite{HSI-Clustering-GCSC-CAI-TGRS-2020,RRCSL-TCYB-WANG-2021}.
Specifically, both metrics range in $[0,1]$, and the higher score
the more accurate clustering result.

\subsection{Quantitative and Qualitative Results}

Table \ref{tab:acc-nmi-small} gives the quantitative comparison of
different clustering models on three small datasets. We can observe
that our FLGC models consistently achieve superior ACC and NMI with
significant margins compared with many existing clustering models.
Specifically, FLGC{*} obtains $0.960$ and $0.983$ ACC on Iris and
Wine datasets, respectively, outperforming the advanced RRCSL ($0.873$
and $0.972$) and KMM ($0.817$ and $0.541$) by large margins. In
Fig. \ref{fig:Clustering-ACC-Bar}, we provide a visual comparison
of the clustering performance on the three challenging datasets. The
results reveal that the proposed FLGC models can markedly improve
many self-expressiveness-based clustering models. As an extension
of subspace clustering, our unsupervised FLGC reduces the intra-class
variations through neighborhood propagation, making it more robust
to find the inherent subspace structure among data. Taking Iris and
Extended Yale B datasets as examples, we visualize the affinity matrices
yielded by FLGC{*}, as shown in Fig. \ref{fig:Affinity}. The visualizations
exhibit distinctly block-diagonal structures, which are highly close
to the corresponding ground truths. 

\subsection{Analysis of Robustness }

To analyze the robustness of FLGCs, we conduct experiments to compare
the clustering performance under different noise conditions. Specifically,
we adopt a Gaussian noise and a salt-and-pepper noise to corrupt images.
The variance $\sigma^{2}$ of the Gaussian noise and the proportion
of corrupted pixels $p$ by the salt and pepper noise are treated
as the intensity of noise. We test our method with different intensities
of 0.01, 0.05, 0.1, and 0.2. It can be observed from Table \ref{tab:robust},
our FLGC{*} is more robust to noise than other methods. This superiority
is benefited from the graph structure, as well as the initial residual
propagation scheme. FLGC and EGCSC have the same performance which
is because FLGC obtains the best performance when $K=1$, and FLGC
degrades into SGCSC at this point. Compared with EDSC, the other three
methods show lower sensitivity to noise, demonstrating the robustness
of the graph convolution. 

\subsection{Influence of $K$ and $\alpha$}

To study the effect of over-smoothing, we show clustering accuracy
under different propagation steps on the COIL20, ORL, and Extended
Yale B datasets, given in Fig. \ref{fig:propagation-clu} (a) and
(b). We find that FLGC{*} (Fig. \ref{fig:propagation-clu} (a)) is
robust to large propagation steps since its performance is almost
unaffected by a large $K$ on the Extended Yale B dataset. In comparison,
the accuracy obtained by FLGC drops about 0.70 in terms of ACC for
$K=20$. This robustness to large propagation step further demonstrates
that the initial features are crucial for regular data. This conclusion
is further supported in Fig. \ref{fig:propagation-clu} (c), where
the clustering ACC tends to be increased by large $\alpha$ for the
ORL and Extended Yale B datasets. It should be noticed that the structure
information shows a higher contribution than initial features for
the COIL20 datasets. A conceivable reason is that the samples within
COIL20 have a strong inter-class difference, which results in a more
accurate structure.

\section{Conclusions\label{sec:Conclusions}}

In this paper, we have presented a unified and simple graph convolutional
framework, i.e., fully linear graph convolution networks, which incorporates
multi-hop neighborhood aggregation into classic linear models to further
simplify the training, applying, and implementing of GCN. Technically,
we train FLGC by computing a global optimal closed-form solution,
resulting in efficient computation. Also, based on the framework,
we developed a semi-supervised FLGC and an unsupervised FLGC for semi-supervised
node classification tasks and unsupervised clustering tasks, respectively.
Furthermore, we showed that FLGC acts as a generalization of traditional
linear models on the non-Euclidean data. In comparison with existing
approaches, our FLGCs achieved superior performance across real-word
graphs and regular grid data concurrently. The success of our FLGC
establishes a connection between GCN and classic linear models. Future
work may include exploring more scalable linear models to deal with
large-scale graph, inductive learning, and extending FLGC to different
GCNs. 

\section*{Acknowlegment}

The authors would like to thank the anonymous reviewers for their
constructive suggestions and criticisms. 

\bibliographystyle{IEEEtran}
\bibliography{Ref-BibTex}

\end{document}